\documentclass[lettersize,journal]{IEEEtran}
\usepackage{amsmath,amsfonts}
\usepackage{algorithmic}
\usepackage{algorithm}
\usepackage{array}
\usepackage{textcomp}
\usepackage{stfloats}
\usepackage{url}
\usepackage{verbatim}
\usepackage{graphicx}
\usepackage{cite}
\usepackage{multirow}
\usepackage{subfigure}
\usepackage{booktabs}
\usepackage{color} 
\usepackage{siunitx}
\usepackage{colortbl}

\definecolor{mygray}{gray}{.9}

\begin{document}

\title{Evaluating Point Cloud from Moving Camera Videos: A No-Reference Metric}

\author{Zicheng Zhang, Wei Sun$^\dag$, Yucheng Zhu, Xiongkuo Min, \emph{Member, IEEE,} \\ Wei Wu, Ying Chen, \emph{Senior Member, IEEE}, and Guangtao Zhai$^\dag$, \emph{Senior Member, IEEE} 
\IEEEcompsocitemizethanks{\IEEEcompsocthanksitem This work was supported in part by NSFC (No.62225112, No.61831015, No. 62301316), the Fundamental Research Funds for the Central Universities, National Key R\&D Program of China 2021YFE0206700, Shanghai Municipal Science and Technology Major Project (2021SHZDZX0102), STCSM 22DZ2229005, and the China Postdoctoral Science Foundation under Grant 2023TQ0212 and 2023M742298. \textit{(Co-corresponding Authors: Wei Sun and Guangtao Zhai.)} \protect}

\IEEEcompsocitemizethanks{\IEEEcompsocthanksitem Zicheng Zhang, Wei Sun, Yucheng Zhu, Xiongkuo Min, and Guangtao Zhai are with the Institute of Image Communication and Network Engineering, Shanghai Jiao Tong University, 200240 Shanghai, China. E-mail:\{zzc1998,sunguwei,zyc420,minxiongkuo,zhaiguangtao\}@sjtu.edu.cn.\protect}
\IEEEcompsocitemizethanks{\IEEEcompsocthanksitem Wei Wu and Ying Chen are with the Alibaba Group, 310052 Hangzhou, China. 
E-mail: \{guokui.ww, chenying.ailab\}@alibaba-inc.com.\protect}}



\maketitle
\begin{abstract}
 Point cloud is one of the most widely used digital representation formats for three-dimensional (3D) contents, the visual quality of which may suffer from noise and geometric shift distortions during the production procedure as well as compression and downsampling distortions during the transmission process. To tackle the challenge of point cloud quality assessment (PCQA), many PCQA methods have been proposed to evaluate the visual quality levels of point clouds by assessing the rendered static 2D projections. Although such projection-based PCQA methods achieve competitive performance with the assistance of mature image quality assessment (IQA) methods, they neglect that the 3D model is also perceived in a dynamic viewing manner, where the viewpoint is continually changed according to the feedback of the rendering device. Therefore, in this paper, we evaluate the point clouds from moving camera videos and explore the way of dealing with PCQA tasks via using video quality assessment (VQA) methods. First, we generate the captured videos by rotating the camera around the point clouds through several circular pathways. Then we extract both spatial and temporal quality-aware features from the selected key frames and the video clips through using trainable 2D-CNN and pre-trained 3D-CNN models respectively. Finally, the visual quality of point clouds is represented by the video quality values.  The experimental results reveal that the proposed method is effective for predicting the visual quality levels of the point clouds and even competitive with full-reference (FR) PCQA methods. The ablation studies further verify the rationality of the proposed framework and confirm the contributions made by the quality-aware features extracted via the dynamic viewing manner.  The code is available at https://github.com/zzc-1998/VQA\_PC.
\end{abstract}

\begin{IEEEkeywords}
point cloud quality assessment, moving camera videos, no-reference, video quality assessment
\end{IEEEkeywords}



\section{Introduction}\label{sec:introduction}
\IEEEPARstart{P}OINT cloud is essentially a huge collection of tiny individual points plotted in three-dimensional (3D) space and each point is represented with the geometry coordinates and may be attached with color or texture. With the rapid development of computer graphics, point cloud has gained a better ability to vividly describe virtual objects and has been widely employed in diverse applications such as virtual reality \cite{xiong2021augmented}, 3D reconstruction \cite{kang2020review}, and metaverse \cite{ning2021survey}. Nevertheless, the visual quality of point clouds can be affected by a variety of distortions. For example, point clouds are usually generated by the 3D laser scanner \cite{chen2013point}, thus disturbances such as noise and slight shifts may be introduced to the constructed point clouds. Moreover, to reduce the storage consumption as well as transmission overload, most point clouds are processed with lossy compression and downsampling, which inevitably damages the perceived quality of point clouds \cite{sim2008compression,park2008multiscale,javaheri2020point,de2018graph}. Therefore, it is necessary to carry out objective point cloud quality assessment (PCQA) metrics to automatically measure the visual quality levels of point clouds.

\begin{figure}
    \centering
    \includegraphics[width = 5cm]{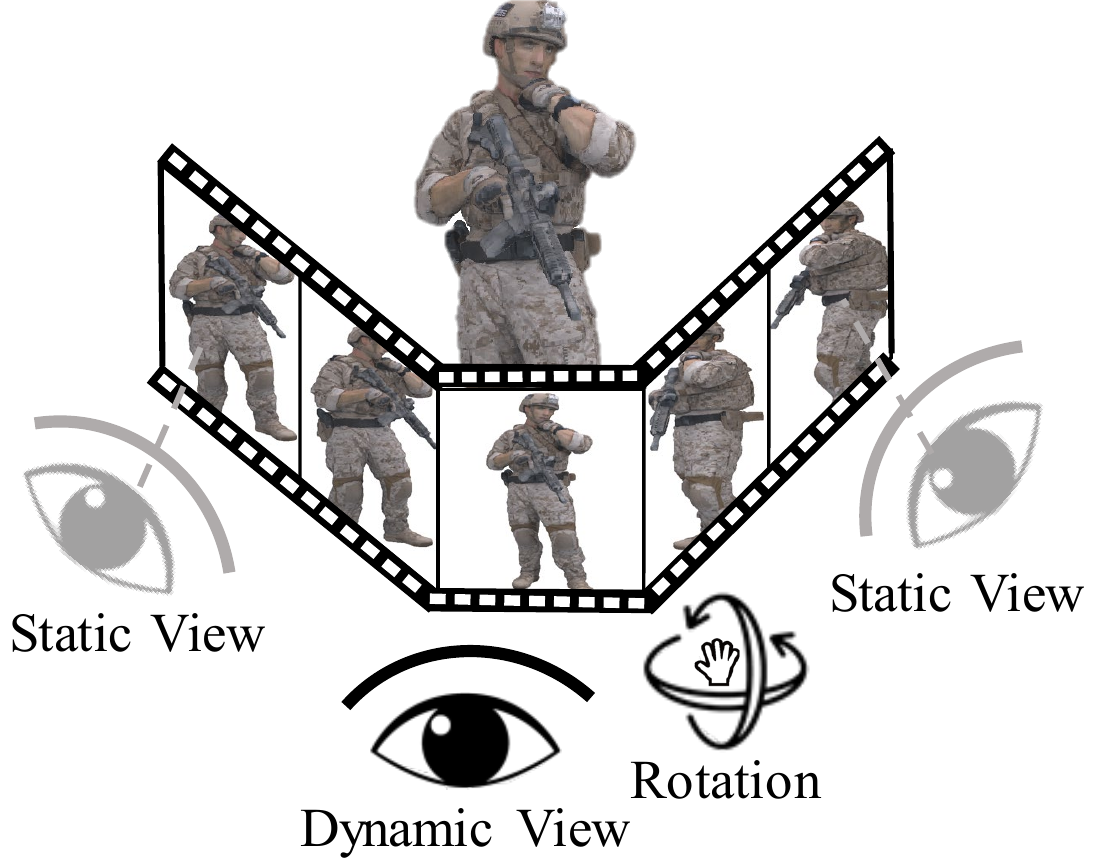}
    \caption{Illustration of human habits of observing point clouds via static and dynamic views. The static views focus on the single rendered images while the dynamic views focus on the frames (usually generated by rotation).}
    \label{fig:view}
    \vspace{-0.5cm}
\end{figure}

In the last decade, large amounts of PCQA methods have been proposed \cite{p2point, p2plane, m1,ff2_roughness, p2mesh,angular,pcqa2,dame,pcqa3,tian-color,guo-color,pcqm,liu2021reduced,zhang2022no,zhang2022mm,zhang2023eep,zhang2023gms}, which can be generally categorized into full-reference (FR), reduced-reference (RR), and no-reference (NR) PCQA methods according to the involve extent of reference information. In many practical situations such as 3D reconstruction, the pristine reference point cloud is not available, thus NR-PCQA methods have wider application, which is also the focus of this paper. Further, NR-PCQA methods can be divided into model-based and projection-based methods. The model-based methods directly take the point cloud as the input to evaluate its quality \cite{p2point,p2plane,p2mesh,alexiou2020pointssim,yang2020graphsim,meynet2020pcqm,zhang2021no} while the projection-based methods assess the quality from the rendered 2D projections \cite{sjtu-pcqa,pcqa_database2,liu2021pqa}. The model-based PCQA methods are more difficult to develop since the point clouds are usually complex and made up of large amounts of points, making it hard to efficiently extract quality-aware features. 
The projection-based PCQA methods in previous works mainly assess the quality by evaluating the 2D views from the separate and static viewpoints \cite{sjtu-pcqa,liu2021pqa}. Such methods are based on the assumption that humans perceive the quality of point clouds through fixed static views. However, this hypothesis does not fully hold in the practical situation. As illustrated in Fig. \ref{fig:view}, humans tend to observe the point cloud through the combination of static and dynamic views. For static views, the viewpoints are fixed and observers simply evaluate the projected images. However, for dynamic views, the viewpoints are continuously adjusted and the observers are just like watching the videos, of which the frames are rendered from each viewpoint. Therefore, it is reasonable to evaluate the visual quality of point clouds by using VQA methods since videos are spatio-temporal media that are able to cover static and dynamic views.  

\begin{figure}[!tp]
    \centering
    \subfigure[]{
    \begin{minipage}[t]{0.32\linewidth}
    \centering
    \includegraphics[width = 2cm, height = 2.5cm]{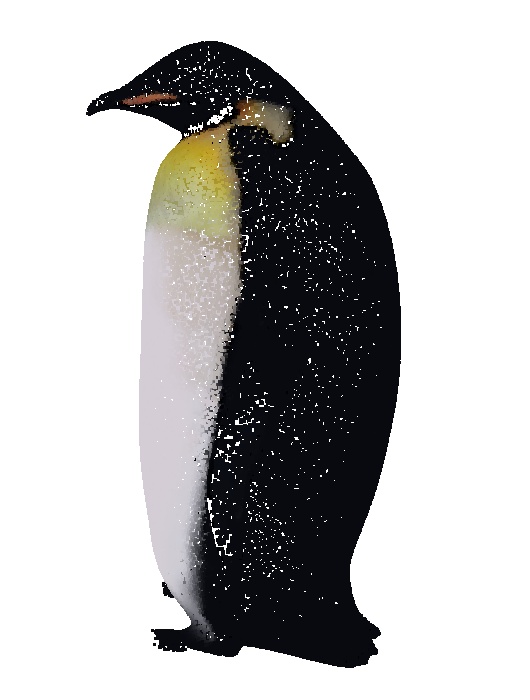}
    \end{minipage}%
    }%
    \subfigure[]{
    \begin{minipage}[t]{0.32\linewidth}
    \centering
    \includegraphics[width = 2cm, height = 2.5cm]{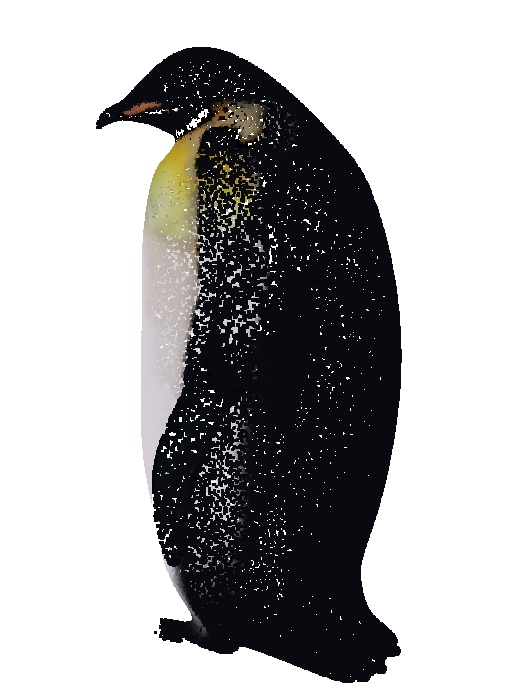}
    \end{minipage}%
    }%
    \subfigure[]{
    \begin{minipage}[t]{0.3\linewidth}
    \centering
    \includegraphics[width = 2cm, height = 2.5cm]{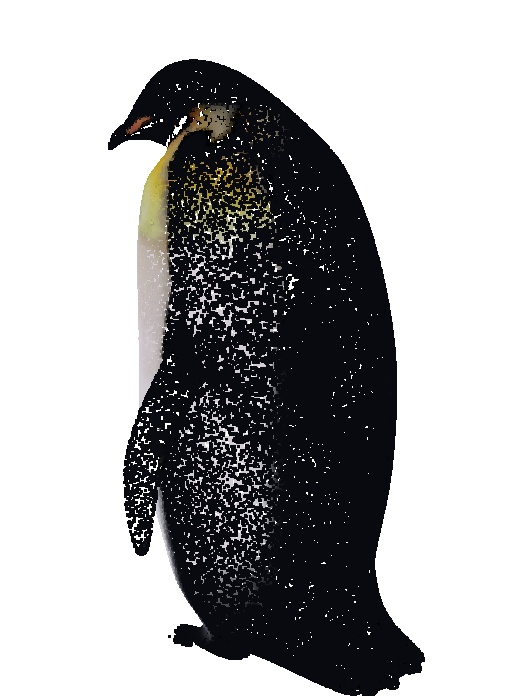}
    \end{minipage}}

     \subfigure[]{
    \begin{minipage}[t]{0.32\linewidth}
    \centering
    \includegraphics[width = 2cm, height = 2.5cm]{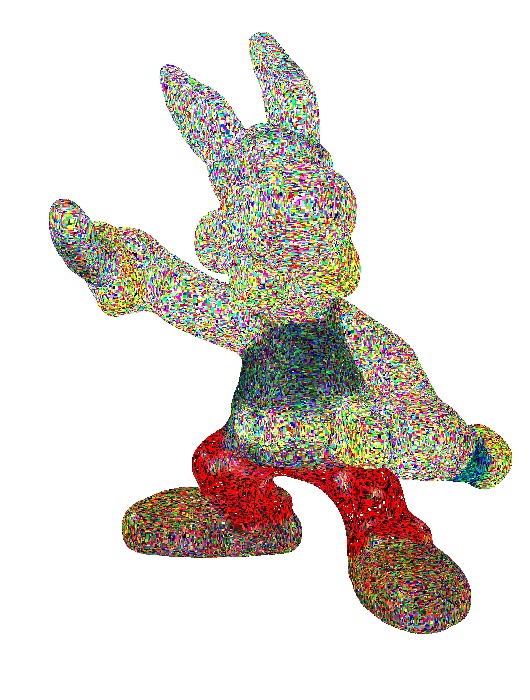}
    \end{minipage}%
    }%
    \subfigure[]{
    \begin{minipage}[t]{0.32\linewidth}
    \centering
    \includegraphics[width = 2cm, height = 2.5cm]{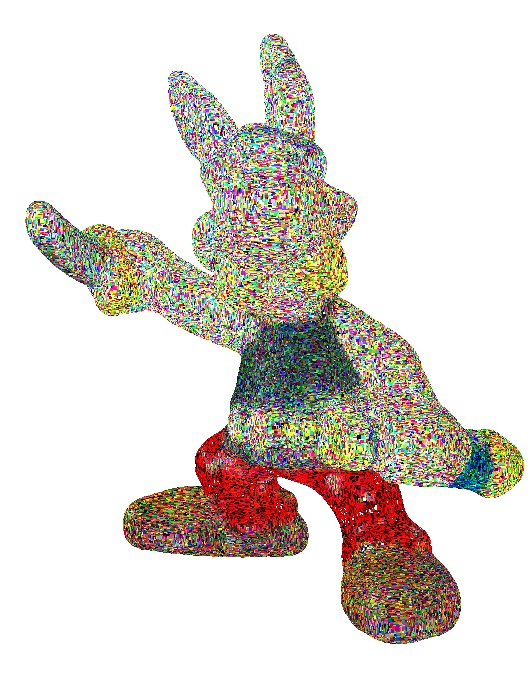}
    \end{minipage}%
    }%
    \subfigure[]{
    \begin{minipage}[t]{0.3\linewidth}
    \centering
    \includegraphics[width = 2cm, height = 2.5cm]{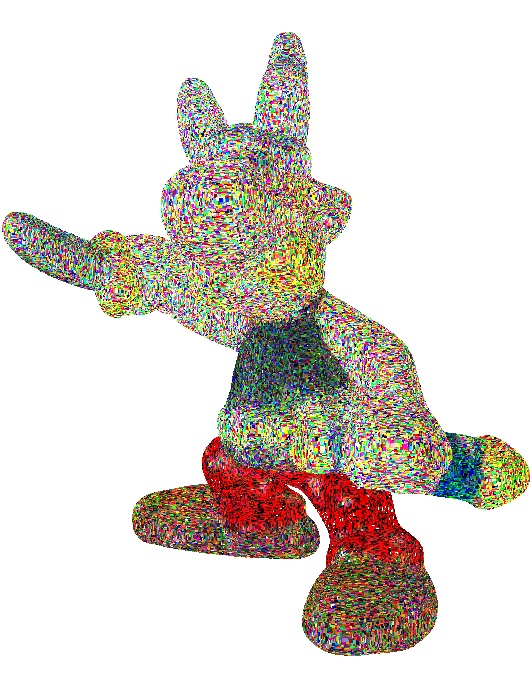}
    \end{minipage}%
    
    }%
    
    \caption{Distortion reflection among dynamic views. (a), (b), and (c) are frames of the point cloud with downsampling distortion while (d), (e), and (f) are frames of the point cloud with color noise.}
    \label{fig:distortion}
\end{figure}

To achieve this end, in this paper, we propose to treat point cloud as moving camera videos and infer the point cloud quality through the quality-aware static (spatial) and dynamic (temporal) information from the video perspective. More specifically, the spatial features of point cloud videos help describe the annoyance of distortions in a static state, while the temporal features of point cloud videos assist to explore the influence of content changes among continuous viewpoints and show whether the view changes are abrupt and unpleasing with certain content and distortions. For example, as shown in Fig. \ref{fig:distortion}, the undesired geometric deformation is obviously abrupt among different viewpoints and the color noise makes the content difference between the adjacent frames more incoherent, even resulting in flickering. Driven by such motivation, we first rotate the camera around the point cloud through four circular pathways to capture four typical video clips to cover sufficient quality-aware content. Each clip has 30 frames and we stitch the four clips into a video consisting of 120 frames. Second, following the strategies in common VQA methods \cite{li2022blindly,sun2022deep}, we extract the spatial and temporal features from selected key frames and video clips by utilizing trainable 2D-CNN and pre-trained 3D-CNN models. Then the obtained spatial and temporal features are processed with linear projection to align the number of channels and then concatenated to form the final quality-aware features. In the last, the features are regressed into quality values through using fully-connected (FC) layers. The experimental results show that the proposed method significantly outperforms the NR-PCQA and is even comparable compared with FR-PCQA methods. The major contributions are summarized as follows:
\begin{itemize}
    \item \textbf{We deal with the PCQA tasks as the VQA problems by evaluating point cloud from moving camera videos}, which pushes forward the development of projection-based PCQA methods by taking advantage of both static and dynamic views.
    \item We design a video capture framework for the PCQA task. The camera is rotated around the point cloud via four symmetric circular pathways to cover sufficient quality-aware content and viewpoint changes.
    \item The proposed method and baseline PCQA methods are validated on the SJTU-PCQA \cite{sjtu-pcqa}, the WPC \cite{liu2021perceptual}, and the LSPCQA-I \cite{pcqa-large-scale} databases. Extensive experimental results show that the proposed method significantly outperforms all compared NR methods and is even comparable with FR methods. Further ablation studies verify the effectiveness of the proposed structure and confirm the improvement made by the dynamic views. 
\end{itemize}

\section{Related Work}
\label{sec:related}
In this section, we give a brief discussion about the development of point cloud quality assessment (PCQA) and video quality assessment (VQA).

\begin{figure*}
    \centering
    \includegraphics[width=18.15cm]{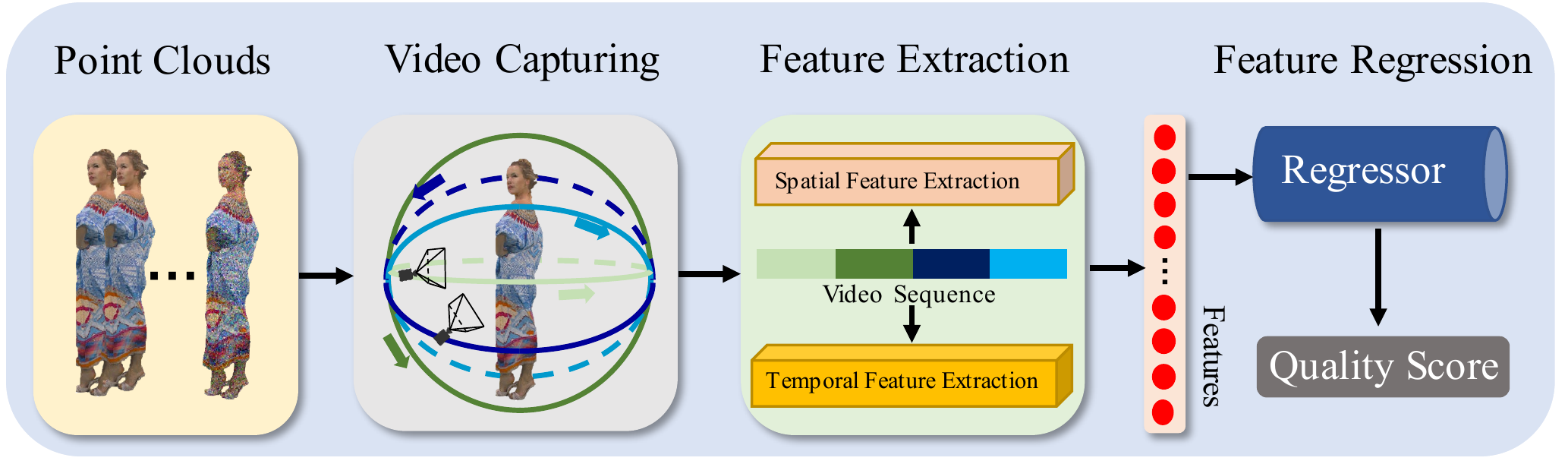}
    \caption{The framework of the proposed method, which includes the video capture process, the feature extraction module, and the feature regression module.}
    \label{fig:framework}
    \vspace{-0.3cm}
\end{figure*}

\begin{figure}
    \centering
    \includegraphics[width = 8.5cm]{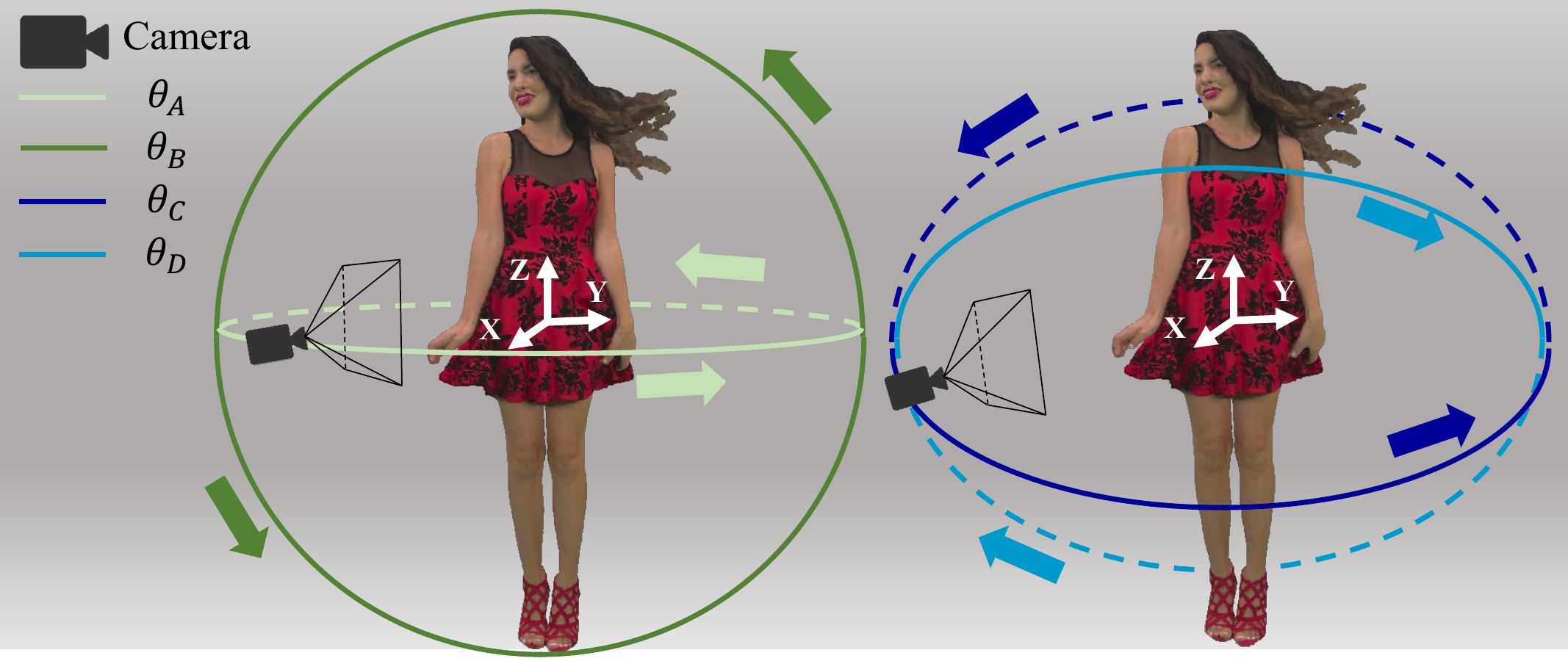}
    \caption{Illustration of video capturing process via moving camera. The point cloud sample is $redandblack$ from the SJTU-PCQA database \cite{sjtu-pcqa}. $\theta_{A}$, $\theta_{B}$, $\theta_{C}$, and $\theta_{D}$ represent the four moving paths respectively.}
    \label{fig:camera}
    \vspace{-0.3cm}
\end{figure}

\subsection{PCQA Development}
The PCQA is still an emerging field in the current literature. The earliest PCQA methods simply focus on the point level, which include p2point \cite{p2point}, p2plane \cite{p2plane}, and p2mesh \cite{p2mesh}, etc. The p2point calculates the distance between corresponding points to infer the distortion levels, the p2plane utilizes the projection of the distance vector along an average normal vector, and the p2mesh computes the distance from points to the reconstructed surface after mesh reconstruction to estimate the quality loss. However, these methods only take geometry information into consideration, thus obtaining unsatisfying performance in the latest colored PCQA databases \cite{sjtu-pcqa,liu2021pqa}. Then PointSSIM \cite{alexiou2020pointssim} predicts the quality difference between the reference and distorted point clouds by comparing the local topology and color distributions. GraphSIM \cite{yang2020graphsim} infers point cloud quality through graph similarity and color gradient. To further take advantage of various features, PCQM \cite{meynet2020pcqm} employs a weighted linear combination of curvature along with color features for evaluation. Besides, some researchers try to evaluate the visual quality of point clouds through 2D projections \cite{sjtu-pcqa,pcqa_database2}, which achieves competitive performance with the assistance of mature IQA methods. { More recently, the multiscale potential energy discrepancy (MPED) \cite{yang2022mped} is introduced for distortion quantification in point clouds, drawing inspiration from classical physics by viewing point clouds as systems with potential energy, which is differentiable, distortion discriminable, and computationally efficient. TCDM \cite{zhang2022evaluating} assesses point cloud quality by gauging the transformational complexity between a distorted cloud and its reference, using space segmentation and a space-aware vector autoregressive model.} 

{ All the PCQA methods mentioned above are FR methods, which require the involvement of reference point clouds, thus having a smaller range of applications. However, limited numbers of NR-PCQA methods have been proposed. ResCNN \cite{pcqa-large-scale} designs an end-to-end sparse convolutional neural network to extract quality-aware features for the point cloud. PQA-net \cite{liu2021pqa} classifies and evaluates the distortions via features extracted from multi-view-based projections. 3D-NSS  \cite{zhang2022no} employs both geometry and color attributes and utilizes some classical statistical distribution models for analysis.  Moreover, IT-PCQA \cite{yang2022no} leverages both subjective scores from natural images and  unsupervised adversarial domain adaptation to infer point cloud quality, which achieves comparable performance. Tu $et$ $al.$ \cite{tu2022v} utilizes a dual-stream convolutional network to evaluate compression distortions in point clouds by extracting both global and local texture and geometry features, and incorporating visual perception of salient regions. GPA-Net \cite{shan2023gpanet} employs a newly proposed graph convolution kernel, GPAConv, to attentively capture point cloud structure/texture perturbations and integrates a multi-task framework addressing quality regression, distortion type, and degree predictions.}

\subsection{VQA Development}
Many VQA methods have been proposed in the last decade \cite{vmaf,sun2023analysis,mittal2015completely,saad2014blind,korhonen2019two,tu2021ugc,saad2014blind,korhonen2019two,tu2021ugc,li2019quality,tu2021rapique,sun2021deep,li2022blindly,sun2022deep,lu2022deep,zhang2023md}, which can be similarly divided into FR-VQA and NR-VQA methods.
FR-VQA methods usually compute the quality difference between the reference and distorted frames with IQA methods (commonly using PSNR, SSIM \cite{ssim}) to represent the video quality. With the development of machine learning, Video Multi-Method Assessment Fusion (VMAF) \cite{vmaf} proposed by Netflix extracts various IQA features along with temporal features and uses Support Vector Regressor for quality regression. 

Similar to the strategy employed in FR-VQA methods, a simple NR-VQA method is to compute each frame's quality level using NR IQA methods (such as NIQE \cite{niqe}) and obtain the video quality via average pooling. To further incorporate spatial with temporal features, many handcrafted-based methods have been proposed to extract spatio-temporal features from the video \cite{mittal2015completely,saad2014blind,korhonen2019two,tu2021ugc}. Namely, VIIDEO \cite{mittal2015completely} quantifies distortion under the intrinsic statistical regularities that are observed in natural videos. V-BLIINDS \cite{saad2014blind} predicts video quality by utilizing a spatio-temporal natural scene statistics model and a motion characterization model. TLVQM \cite{korhonen2019two} calculates low complexity and high complexity features in two steps for video quality evaluation. VIDEVAL \cite{tu2021ugc} selects representative quality-aware features among the top NR-VQA models and fuses the  features into quality scores. Considering the huge success of deep learning, deep neural networks (DNN) have been employed for VQA tasks as well. VSFA \cite{li2019quality} extracts deep semantic features with a pre-trained DNN model and models the temporal effect with gated recurrent units (GRUs). RAPIQUE \cite{tu2021rapique} combines and leverages the advantages of both quality-aware scene statistics features and semantics-aware deep features for video quality prediction. StairVQA \cite{sun2021deep} further utilizes both low-level and high-level visual features from the deep neural network as the quality-aware feature representation. To better understand video distortions caused by the motion, some studies \cite{li2022blindly,sun2022deep} attempt to extract temporal features with 3D-CNN models pre-trained on the video action recognition databases to enhance video quality understanding and have achieved strong performance.

\section{Proposed Method}
\label{sec:proposed}
{ The framework of the proposed method is clearly exhibited in Fig. \ref{fig:framework}, which includes the video capture module, the feature extraction module, and the feature regression module.}

\begin{figure*}
    \centering
    \subfigure[]{\includegraphics[width = 18 cm]{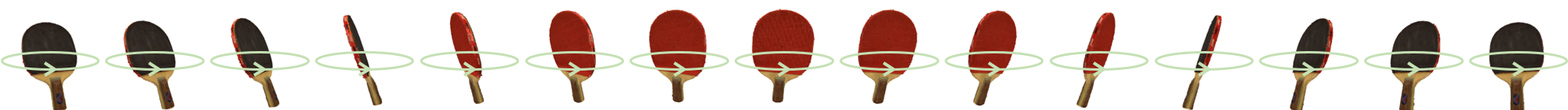}}
    \subfigure[]{\includegraphics[width = 18 cm]{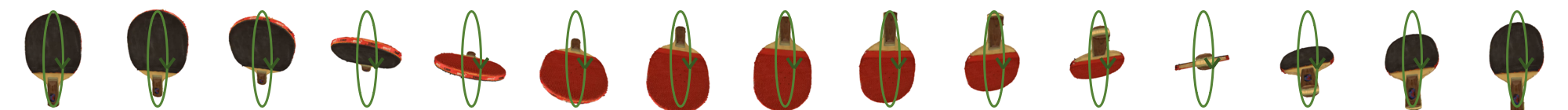}}
    \subfigure[]{\includegraphics[width = 18 cm]{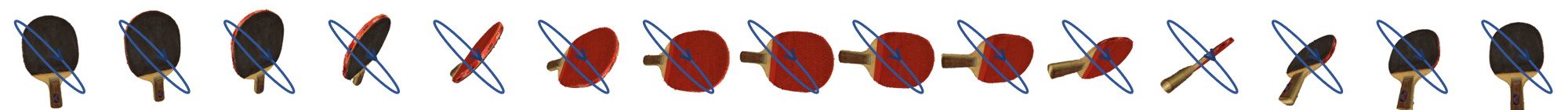}}
    \subfigure[]{\includegraphics[width = 18 cm]{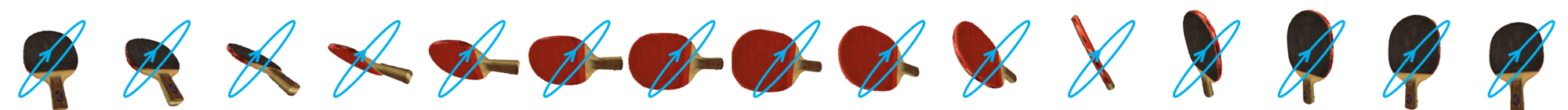}}
    \caption{Example of the captured video from the point cloud $ping\_pong\_bat$ from the WPC database \cite{liu2021perceptual}, where (a), (b), and (c) represent the samples frames captured of ${\theta_{A}}$, ${\theta_{B}}$, ${\theta_{C}}$, and ${\theta_{D}}$ respectively.}
    \label{fig:frames}
    \vspace{-0.3cm}
\end{figure*}

\subsection{Video Capture}
\subsubsection{Pathways}
We assume that humans evaluate the visual quality of point clouds through both static and dynamic views. Static views can be modeled by the projections from fixed viewpoints while the dynamic views can be modeled by the videos, of which the frames are rendered from continuously changing viewpoints. 
 Given a point cloud $\mathbf{P} = \{\delta_{m} , \gamma_{m} \}_{m=1}^{N}$, where $\delta_{m} \in \mathbb{R}^{1 \times 3} $ indicates the geometry coordinates, $\gamma_{m} \in \mathbb{R}^{1 \times 3} $ represents the attached RGB color information, and $N$ stands for the number of points. The corresponding video sequence $\mathbf{V}$ is generated with the assistance of the Python package open3d \cite{Zhou2018}:
\begin{equation}
    \mathbf{V} = \Psi(\mathbf{P}),
\end{equation}
where $\Psi(\cdot)$ represents the video capture operation. Specifically, the camera is first placed at the default position with a fixed viewing distance $R$. Then we define the geometry mean center of the point cloud $\mathbf{P}$ as:
\begin{align}
\begin{aligned}
O_{\delta} &  = \frac{1}{N} \sum_{m=1}^{N} \delta_{m}, 
\end{aligned}
\end{align}
where the $O_{\delta}$ represents the ($X,Y,Z$) coordinates of the point cloud's mean center, and $\delta_{m}$ stands for the ($X,Y,Z$) coordinates of the $m$-th point in the point cloud. Next, we establish a space Cartesian coordinate system with $O_{\delta}$ as the center and the position coordinates of the moving camera are defined as ($X_{\alpha},Y_{\alpha},Z_{\alpha}$). To cover as many viewpoints as possible, we especially design 4 symmetric circular rotation pathways to model the temporal features of the point cloud, which can be represented as:
\begin{equation}
\theta_{A}:\left\{
\begin{aligned}
X_{\alpha}^2 + Y_{\alpha}^2 &= R^2, \\
Z_{\alpha} &= 0, 
\end{aligned}
\right.
\end{equation}
\begin{equation}
\theta_{B}:\left\{
\begin{aligned}
Y_{\alpha}^2 + Z_{\alpha}^2 &= R^2, \\
X_{\alpha} &= 0, 
\end{aligned}
\right.
\end{equation}
\begin{equation}
\quad \quad  \quad {\theta_{C}:}\left\{
\begin{aligned}
X_{\alpha}^2 + Y_{\alpha}^2 + Z_{\alpha}^2 &= R^2, \\
X_{\alpha} + Z_{\alpha} &= 0, 
\end{aligned}
\right.
\end{equation}
\begin{equation}
\quad \quad  \quad {\theta_{D}:}\left\{
\begin{aligned}
X_{\alpha}^2 + Y_{\alpha}^2 + Z_{\alpha}^2 &= R^2, \\
X_{\alpha} - Z_{\alpha} &= 0, 
\end{aligned}
\right.
\end{equation}
where ${\theta_{A}}$, ${\theta_{B}}$, ${\theta_{C}}$, and ${\theta_{D}}$ represent the four corresponding pathways and $R$ indicates the radius of the circle. The illustration of the video capture process is clearly shown in Fig. \ref{fig:camera}. { Paths \( \theta_A \) and \( \theta_B \) represent 2D circles with radius $R$ that lie in the XY and YZ planes. Paths \( \theta_C \) and \( \theta_D \) are defined to represent circles on the surface of a sphere bisected by the diagonal planes. The tilt of 45° relative to both the XY and XZ planes in paths \( \theta_C \) and \( \theta_D \) provides a symmetric and simple perspective between the primary planes, which can maintain uniformity relative to the primary XY and XZ planes, offering a balanced view between them. }

\subsubsection{Camera Rotation}
To maintain the consistency of the videos, the camera rotation step is set as 12$^\circ$ between consecutive frames through trial. To be more specific, the camera rotates 12$^\circ$ around the center of the circular pathway to capture the next frame after capturing the current frame. Then a clip of 360/12 = 30 frames is captured to cover each circular pathway, and a video consisting of four clips and 4$\times$30 = 120 frames is obtained for the point cloud. In addition, the camera returns to the same starting position after sampling frames for each pathway, therefore, the captured video is continuous among the clips. An example of the captured video is presented in Fig. \ref{fig:frames}, from which we can distinctly observe the rotation changes reflected through the four pathways.


\subsection{Feature Extraction}

\begin{figure}
    \centering
    \includegraphics[width=8.7cm]{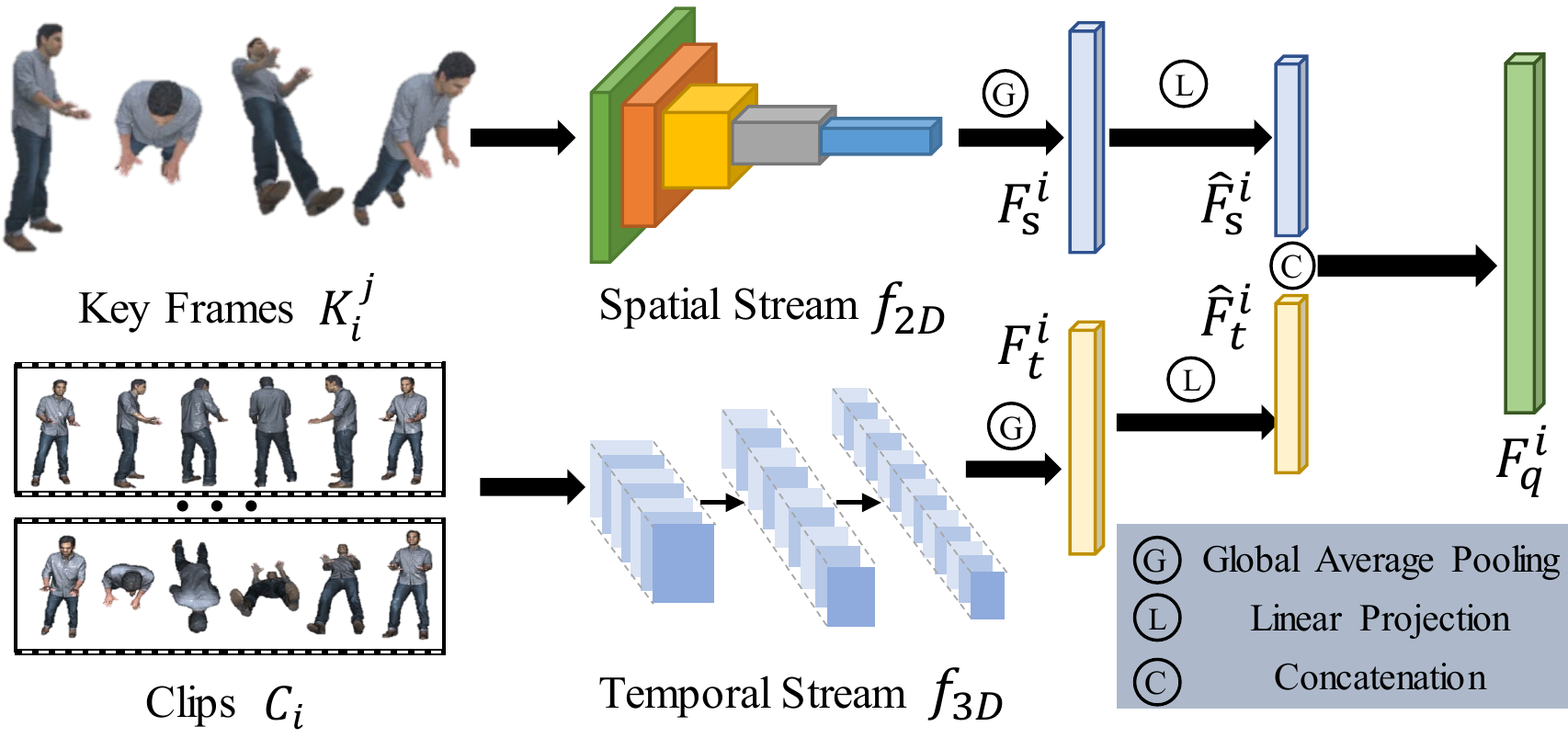}
    \caption{The detailed structure of the feature extraction.}
    \vspace{-0.2cm}
    \label{fig:extraction}
\end{figure}

Similar to most VQA methods, we try to extract the quality-aware information of the video clips from the spatial and temporal domains. Therefore, inspired by previous VQA models \cite{li2022blindly,sun2022deep}, we utilize a 2D and 3D combined approach for feature extraction, which is illustrated in Fig. \ref{fig:extraction}.
Following the video capture process described above, we can obtain a video sequence $\mathbf{V}$ with four clips $\{{C_{i}}\}_{i=1}^4$ according to the four pathways and each clip is composed of 30 frames. Then a key frame $K_{i}^j$ ($j$ is the frame index within the clip, 0 $\leq$ $j$  \textless 30) chosen in each single clip $C_{i}$ is selected for spatial feature extraction while each whole single clip $C_{i}$ is employed for the temporal feature extraction respectively.

\subsubsection{Spatial Feature Extraction}
{ The key frame is selected with the viewpoint-max-distance (VMD) strategy, which indicates that the frames with the farthest viewpoints between each other are chosen:
\begin{equation}
K_{i}^j \longleftarrow  \underset{\substack{1 \leq j_1, j_2 \leq 4 \\ j_1 \neq j_2}}{\max} d(VP_{i}^{j_1}, VP_{i}^{j_2}) ,
\end{equation}
where $VP_{i}^{j}$ indicates the viewpoint of sampled key frames from the $j$-th clip and $d(\cdot)$ indicates the distance function. Moreover, if there exist multiple key frame selections satisfying the max distance requirement, the first key frame selection is employed for simplification. As a result, the frames we sample encapsulate a more comprehensive portion of the point cloud content.}
Given the sampled key frame $K_{i}^j$, we simply use a trainable 2D-CNN model to extract the spatial feature maps:
\begin{equation}
\begin{aligned}
F_{s}^i &= f_{2D}(K_{i}^j),\\
\hat{ F_{s}^i} &= {\rm {GAP}}(F_{s}^i), 
\end{aligned}
\end{equation}
where $f_{2D}(\cdot)$ denotes the 2D-CNN model, $F_{s}^i \in \mathbb{R}^{C \times H \times W}$ represents the spatial feature maps for $K_{i}^j$, ${\rm {GAP}}(\cdot)$ indicates the global average pooling function, and $\hat{F}_{s}^i \in \mathbb{R}^{C_{s} \times 1}$ stands for the averaged spatial feature representation.

\subsubsection{Temporal Feature Extraction}
Unlike the spatial features focusing on single frames, temporal features pay more attention to the relations between continuous frames, which are therefore more capable of reflecting the perceptual distortions in the temporal domain. To extract the temporal features, we employ the pre-trained 3D-CNN model as the temporal feature extractor, which has been a common approach in many VQA studies \cite{li2022blindly,sun2022deep}. Moreover, the feature representation acquired by the pre-trained 3D-CNN model is believed to have a strong ability to capture the effect of content changes and model the motion information that is highly correlated with the human vision system (HVS) \cite{li2022blindly}. In addition, the temporal features are not sensitive to the resolution and we downsample the clips to a low resolution before temporal feature extraction. Given the clip $C_{i}$ and the pre-trained 3D-CNN model, the temporal features can be derived as:
\begin{align}
\begin{aligned}
F_{t}^i &= f_{3D}(C_{i}), \\
\hat{ F_{t}^i} &= {\rm {GAP}}(F_{t}^i), 
\end{aligned}
\end{align}
where $f_{3D}(\cdot)$ stands for the pre-trained 3D-CNN model, $\hat{ F_{t}^i} \in \mathbb{R}^{C_{t}\times 1}$ is the temporal quality-aware feature representation after global averaging pooling.

\subsubsection{Feature Fusion}
Since the number of output channels of spatial features and temporal features usually differ from each other,  the linear projection is utilized to align the number of channels. Then the spatial and temporal features are concatenated to form the final quality-aware features:
\begin{equation}
 F_{q}^{i} = W_{s}\hat{F_{s}^i} \oplus W_{p}\hat{ F_{t}^i},   
\end{equation}
where $W_{s}$ and $W_{p}$ are learnable linear mapping matrices, $W_{s}\hat{F_{s}^i}$ and $W_{p}\hat{ F_{t}^i} \in \mathbb{R}^{C'\times 1 }$ ($C'$ is the number of aligned channels), $\oplus$ indicates the concatenation operation, $F_{q}^{i} \in \mathbb{R}^{2C'\times 1 } $ 
represents the combined quality-aware feature for clip $C_{i}$ .

To sum up, we select a key frame $K_{i}^j$ from the clip $C_{i}$ for the spatial feature extraction and we use the whole clip $C_{i}$ for the temporal feature extraction. Finally, the quality-aware features $F_{q}^{i}$ for clip $C_{i}$ are obtained via the linear projection and concatenation. 

\subsection{Feature Regression}
With the generated quality-aware features, we employ a two stage fully-connected (FC) layers consisting of 128 and 1 neuron respectively for feature regression, which can be derived as:
\begin{equation}
    Q_{i} = {\rm FC}(F_{q}^{i}),
\end{equation}
where ${\rm FC}(\cdot)$ denotes the FC layers and $Q_{i}$ stands for the quality score for the $i$-th clip $C_{i}$. Considering that each video includes four clips, we generate the overall quality score through average pooling:
\begin{equation}
    Q = \frac{1}{4} \sum_{i=1}^{4} Q_{i},
\end{equation}
where $Q$ represents the predicted quality level for the video. 

\subsection{Loss Function}
To directly investigate the effectiveness of the VQA application for the point cloud, we directly use the Mean Squared Error (MSE) as the loss function:
\begin{equation}
    Loss = \frac{1}{n}\sum_{\eta=1}^{n}\left(L_{\eta}-Q_{\eta}\right)^{2}
\end{equation}
where $n$ indicates the number of videos in a mini-batch, $L_{\eta}$ and $Q_{\eta}$ are the subjective quality labels and predicted quality levels respectively.

\section{Experiment}
\label{sec:experiment}
In this section, we give the details of the experimental setup. We select several state-of-the-art PCQA and VQA methods for comparison. Further ablation studies are carried out to verify the effectiveness of the proposed method. 
\subsection{Experiment Setup}
\subsubsection{Benchmark Databases}
We mainly validate the effectiveness of the proposed method on the subjective point cloud assessment database (SJTU-PCQA) \cite{sjtu-pcqa}, the Waterloo point cloud assessment database (WPC) \cite{liu2021perceptual}, and the large-scale point cloud quality assessment database part I (LSPCQA-I) \cite{pcqa-large-scale}. The SJTU-PCQA database provides 10 groups of distorted point clouds generated from 10 reference point clouds and 9 groups of point clouds are publicly available. Each group contains 42 distorted point clouds obtained from seven kinds of synthetic distortions with six strengths, which generates 9$\times$42 = 378 distorted point clouds in total. The WPC database includes 740 distorted point clouds generated from 20 reference point clouds by manually introducing downsampling, Gaussian noise, and three types of compression distortions. The LSPCQA-I database is made up of 930 distorted point clouds generated with 5 distortion levels of 31 types of distortions. More complex distortions such as contrast distortion, Gamma noise, mean shift, local missing, etc. are introduced to this database, which makes the LSCPQA-I database more challenging. 

\subsubsection{Comparing Models}
The comparing PCQA methods can be divided into two categories:
\begin{itemize}
    \item  Model-based methods: The methods assess the visual quality of distorted point clouds directly from the 3D digital representation. The FR model-based methods include GraphSIM \cite{yang2020inferring}, PointSSIM \cite{pcqa3}, and PCQM \cite{pcqm}. The RR model-based methods include PCMRR \cite{viola2020reduced}. { The NR model-based method includes 3D-NSS \cite{zhang2022no} and ResSCNN \cite{pcqa-large-scale}.}

    \item Projection-based methods: The projection-based methods evaluate the visual quality of point clouds via the rendered 2D projections. The FR projection-based methods consist of PSNR and SSIM \cite{ssim}. The NR projection-based methods include BRISQUE \cite{brisque}, NIQE \cite{niqe}, PQA-net \cite{liu2021pqa}, VIIDEO \cite{mittal2015completely},V-BLIINDS \cite{saad2014blind}, TLVQM \cite{korhonen2019two}, VIDEVAL \cite{tu2021ugc}, VSFA \cite{li2019quality}, RAPIQUE \cite{tu2021rapique}, StairVQA \cite{sun2021deep}, {BVQA \cite{li2022blindly}, and SimpVQA \cite{sun2022deep}. It's worth mentioning that PSNR, SSIM, NIQE, and BRISQUE are IQA methods and we validate these methods by averaging the frames' quality scores. The VIIDEO, V-BLIINDS, TLVQM, VIDEVAL, VSFA, RAPIQUE, StairVQA, BVQA, and SimpVQA are VQA methods, among which VIIDEO, V-BLIINDS, TLVQM, and VIDEVAL are handcrafted-based methods while the rest is DNN-based methods.} 
\end{itemize}

We employ the $k$-fold cross validation strategy for experiment as suggested in \cite{chetouani2021deep,fan2022no}.
For the SJTU-PCQA database, we select 8 groups of point clouds for training and leave 1 group for testing. Such split process is repeated 9 times so that each group of point cloud is exactly tested only once. For the WPC database, we divide the 20 groups of point clouds into 5 folds and each fold consists of 4 groups of point clouds. Then the similar testing process is utilized. For the LSPCQA-I database, the same 5 fold cross validation is employed as well. The average performance of the $k$ folds is recorded as the final experimental results. It's worth noting that there is no content overlap between the train and test sets. We strictly retrain the compared NR-PCQA methods utilizing the same train and test sets splits. What's more, for the FR-PCQA and RR-PCQA methods that require no training, we simply validate these methods on the same test sets and report the average performance to make the comparison fair.


\subsubsection{Evaluation Criteria}
We employ four evaluation criteria to judge the correlation between the predicted scores and MOSs, which consist of Spearman Rank Correlation Coefficient (SRCC), Kendall’s Rank Correlation Coefficient (KRCC), Pearson Linear Correlation Coefficient (PLCC), and Root Mean Squared Error (RMSE).
An excellent model should get SRCC, KRCC, and PLCC values as close as possible to 1, and get the RMSE value as close as possible to 0.

Additionally, to deal with the scale differences between the predicted quality scores and the quality labels among different PCQA methods, a five-parameter logistic function is employed to map the predicted scores before computing the criteria values:
\begin{equation}
\hat{y}=\beta_{1}\left(0.5-\frac{1}{1+e^{\beta_{2}\left(y-\beta_{3}\right)}}\right)+\beta_{4} y+\beta_{5},
\end{equation}
where $\left\{\beta_{i} \mid i=1,2, \ldots, 5\right\}$ are parameters to be fitted, $y$ and $\hat{y}$ are the predicted scores and mapped scores respectively.

\subsubsection{Implementation Details}
Specifically, the ResNet50 \cite{he2016deep} is used as the spatial feature extractor. 
The key frame index is set as 7, which indicates that the 7-th frame of each clip is used for spatial feature extraction. Since the spatial features are sensitive to the resolution, we maintain the original resolution (1920$\times$1080$\times$3) of the key frames. Patches with the resolution of 224$\times$224$\times$3 are randomly cropped from the key frames for training while patches with the same resolution are cropped in the center for testing. The SlowFast R50 \cite{feichtenhofer2019slowfast} is utilized as the temporal feature extractor and only the fast pathways are used. The clips are resized to 224$\times$224 for both training and testing. Additionally, the ResNet50 is initialized with the pre-trained model on the ImageNet database \cite{deng2009imagenet}
and is fine-tuned during the training stage. The SlowFast R50 is frozen with the weights of the pre-trained model on the Kinetics 400 database \cite{kay2017kinetics}.
The adam optimizer \cite{kingma2014adam} is employed. The default batch size is set as 32 and the default training epochs are set as 50. The initial learning rate is set as 5e-5, which decays with a ratio of 0.9 for every 10 epochs.

\begin{table*}[!tp]
\centering 
\renewcommand\tabcolsep{2.3pt}
\caption{Performance results on the SJTU-PCQA, WPC, and LSPCQA-I databases. The best performance results are marked in {\bf\textcolor{red}{RED}} and the second performance results are marked in {\bf\textcolor{blue}{BLUE}}.}
\begin{tabular}{c|c|c|c|cccc|cccc|cccc}
\toprule
\multirow{2}{*}{Ref}&\multirow{2}{*}{Type}&\multirow{2}{*}{Index}&\multirow{2}{*}{Methods} & \multicolumn{4}{c|}{SJTU-PCQA} & \multicolumn{4}{c|}{WPC} & \multicolumn{4}{c}{LSPCQA-I} \\ \cline{5-16}
        &&& & SRCC$\uparrow$      & PLCC$\uparrow$      & KRCC$\uparrow$     & RMSE $\downarrow$    & SRCC$\uparrow$      & PLCC$\uparrow$      & KRCC$\uparrow$       & RMSE $\downarrow$ & SRCC$\uparrow$      & PLCC$\uparrow$      & KRCC $\uparrow$      & RMSE $\downarrow$\\ \hline
\multirow{5}{*}{FR} &\multirow{3}{*}{Model-based} 
 &A& PCQM        & \bf\textcolor{blue}{0.8644}   & \bf\textcolor{red}{0.8853}    & \bf\textcolor{red}{0.7086}     & \bf\textcolor{blue}{1.0862}     & {0.7434}    & {0.7499}   & \bf\textcolor{blue}{0.5601}   & 15.1639  &0.3911 & 0.4044 & 0.3119 & 0.7527           \\
&&B& GraphSIM    & \bf\textcolor{red}{0.8783}    & 0.8449    & \bf\textcolor{blue}{0.6947}   & \bf\textcolor{red}{1.0321}  & 0.5831    & 0.6163    & 0.4194   & 17.1939  &0.3112 & 0.3774 & 0.2166 & 0.7426 \\
&&C& PointSSIM      & 0.6867  & 0.7136  & 0.4964 & 1.7001  & 0.4542    & 0.4667    & 0.3278   & 20.2733 & 0.3155 &0.5346 &0.2443 & 0.7564 \\\cline{2-16}
&\multirow{2}{*}{Projection-based} &D&
PSNR &0.2952 &0.3222 &0.2048 &2.2972  &0.1261 &0.1801 &0.0897 &22.5482 &0.5580 &0.5909 &0.3989 &0.6640\\
&&E&SSIM &0.3850 &0.4131 &0.2630 &2.2099  &0.2393 &0.2881 &0.1738 &21.9508 &0.5564 & 0.5580 &0.3815 & 0.6129\\ 
 \hline
\multirow{1}{*}{RR} &Model-based&
 F&PCMRR      & 0.4816  & 0.6101  & 0.3362 & 1.9342 & 0.3097    & 0.3433    & 0.2082   & 21.5302 &0.1673 &0.1650 &0.1153 &0.8086\\ \hline
\multirow{11}{*}{NR} & \multirow{2}{*}{Model-based} 
&G&3D-NSS       & 0.7144 & 0.7382  & 0.5174 & 1.7686    & 0.6479    & 0.6514    & 0.4417   & 16.5716 &0.4509 &0.4751&0.3180&0.7317\\
&&H&ResSCNN   & 0.7911 & 0.7821  & 0.5224 & 1.3651    & 0.5533    & 0.5466    & 0.3417   & 18.0770 &0.5611 &0.5531 &0.4376 &0.6929\\\cline{2-16} &\multirow{13}{*}{Projection-based}
&I&BRISQUE  & 0.3975    & 0.4214  & 0.2966 & 2.0937  & 0.2614    & 0.3155  & 0.2088 & 21.1736 &0.2910 &0.2951 &0.1227 &0.8038\\
&&J&NIQE &0.1379 &0.2420 &0.1009 &2.2622  &0.1136 & 0.2225 &0.0953 &23.1415 &0.1127 & 0.2577 & 0.0797 & 0.7982\\
&&K&PQA-net   & 0.8372   & 0.8586    & 0.6304 & {1.0719}  & 0.7034    & 0.7113    & 0.4956   & 15.0210  &0.5711 &0.5616 & 0.4127 &0.6181  \\
&&L&VIIDEO &0.0509 &0.2960 &0.0433 &2.3180  &0.0775 &0.0823  &0.0538 &22.9231 &0.0733 &0.2066 &0.0511 & 0.8030\\
&&M&V-BLIINDS &0.6804 & 0.7822 &0.4863 &1.5091  &0.4618 &0.4903 &0.3096 &19.7348 &0.2008 &0.2439 &0.1372 &0.8071 \\
&&N&TLVQM &0.5270 &0.6081 &0.3442 &1.9125  &0.0366 & 0.0191 &0.2081 &22.1481&0.3381&0.3997&0.2324&0.7628 \\
&&O&VIDEVAL &0.6027 & 0.7465 &0.4259 & 1.5079  &0.3744 & 0.2653 &0.3630 &21.0992 &0.3676 &0.4249&0.2575&0.7533 \\
&&P&VSFA  & 0.7233    &0.8230 & 0.5477  &  1.4096 & 0.6355   & 0.6331    & 0.4649 & 17.2344 & 0.5557 &0.5659 &0.4124 & 0.6542\\
&&Q&RAPIQUE &0.4441 & 0.4054 &0.3417 &2.2144  &0.2793 & 0.3577 &0.2062 &21.1455 &0.3787&0.4130&0.2624&0.7579 \\
&&R&StairVQA  & 0.7812 & 0.7744 & 0.5635 & 1.4564 & 0.7314 &0.7239 & 0.5584 & \bf\textcolor{blue}{14.1723} & {0.6013} & {0.6243} & {0.4508} & {0.6109}\\
&&S&BVQA  & 0.7644 & 0.7543 & 0.5109 & 1.6688 & 0.6920 &0.6881 & 0.4925 & {17.0872} & {0.5017} & {0.4919} & {0.3001} & {0.6670}\\
&&T&SimpVQA  & 0.8200 & 0.8332 & 0.6331 & 1.2779 & \bf\textcolor{blue}{0.7712} & \bf\textcolor{blue}{0.7755} & 0.5449 & {14.2134} & \bf\textcolor{blue}{0.6611} & \bf\textcolor{blue}{0.6724} & \bf\textcolor{blue}{0.4889} & \bf\textcolor{blue}{0.6003}\\
&&U& Proposed   & {0.8611}   &  \bf\textcolor{blue}{0.8702}   & 0.6811 & 1.1012  & \bf\textcolor{red}{0.8012}    & \bf\textcolor{red}{0.8001}    & \bf\textcolor{red}{0.6237}  & \bf\textcolor{red}{13.5578} & \bf\textcolor{red}{0.7056} & \bf\textcolor{red}{0.7244} & \bf\textcolor{red}{0.5331} & \bf\textcolor{red}{0.5561}\\

                      \bottomrule
\end{tabular}
\label{tab:experiment}
\vspace{-0.3cm}
\end{table*}

\begin{figure}[!tp]
    \centering
    \subfigure[Sample NSI]{\includegraphics[width = 3.8cm]{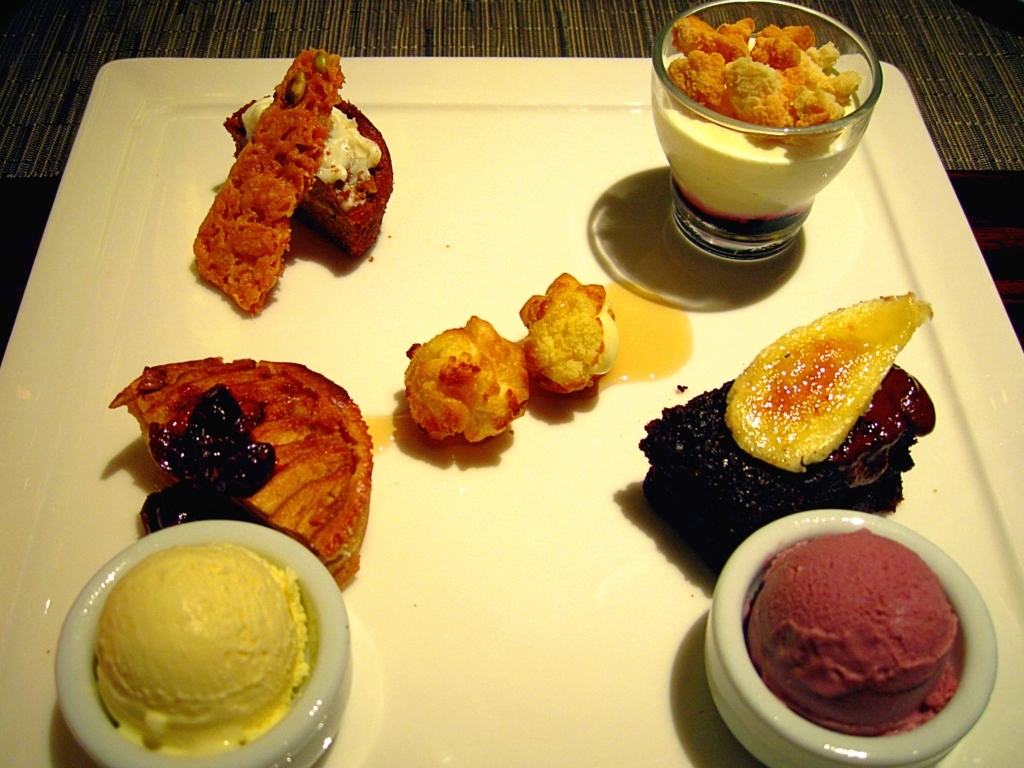}}\hspace{0.18cm}
    \subfigure[Sample PCI]{\includegraphics[width = 3.8cm ]{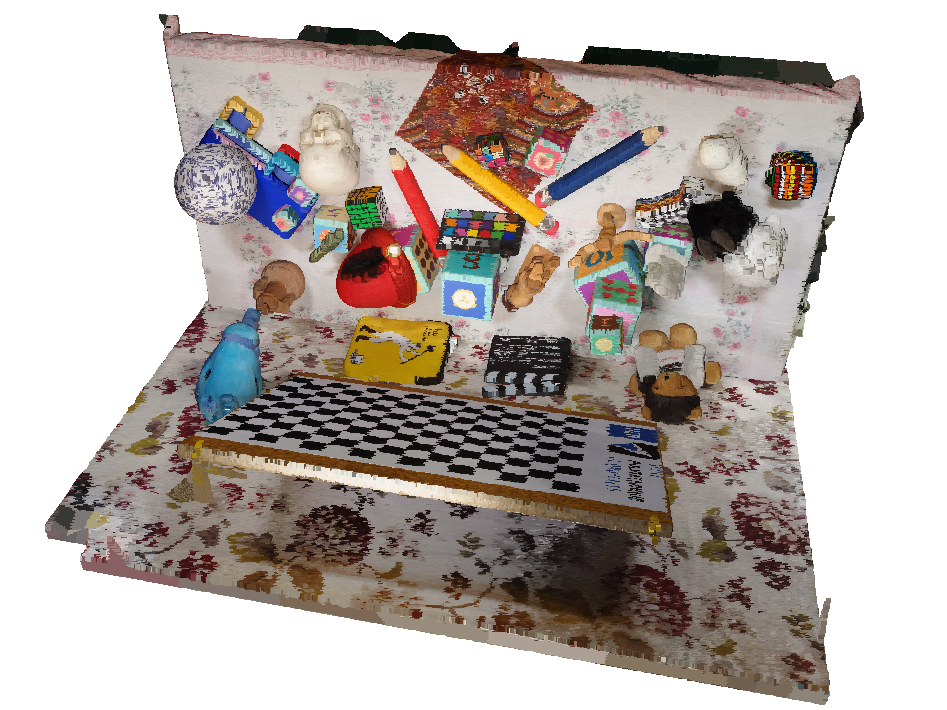}}
    \subfigure[NSI distributions]{\includegraphics[width = 4cm,height = 2.5cm]{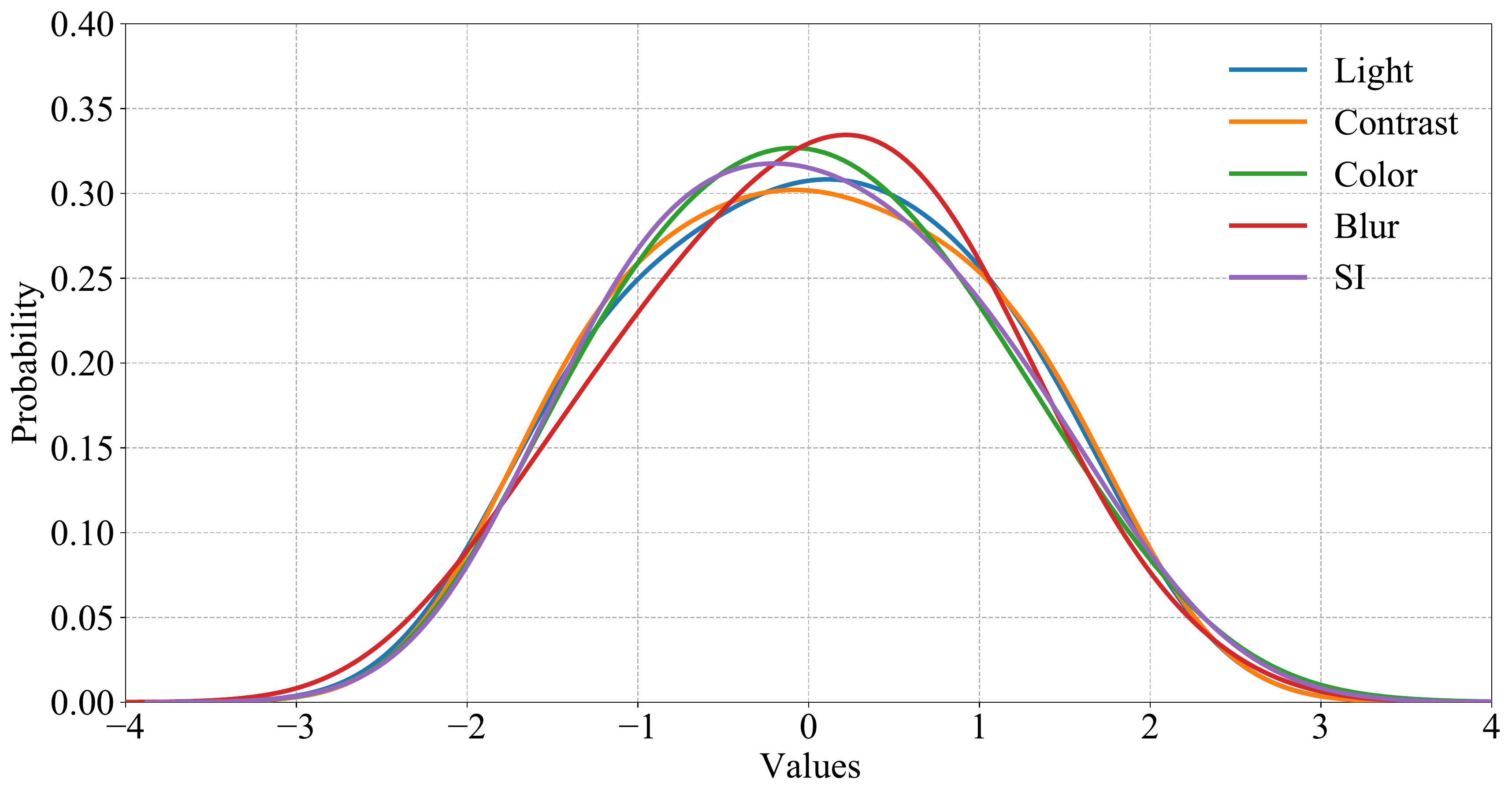}}
    \subfigure[PCI distributions]{\includegraphics[width = 4cm,height = 2.5cm]{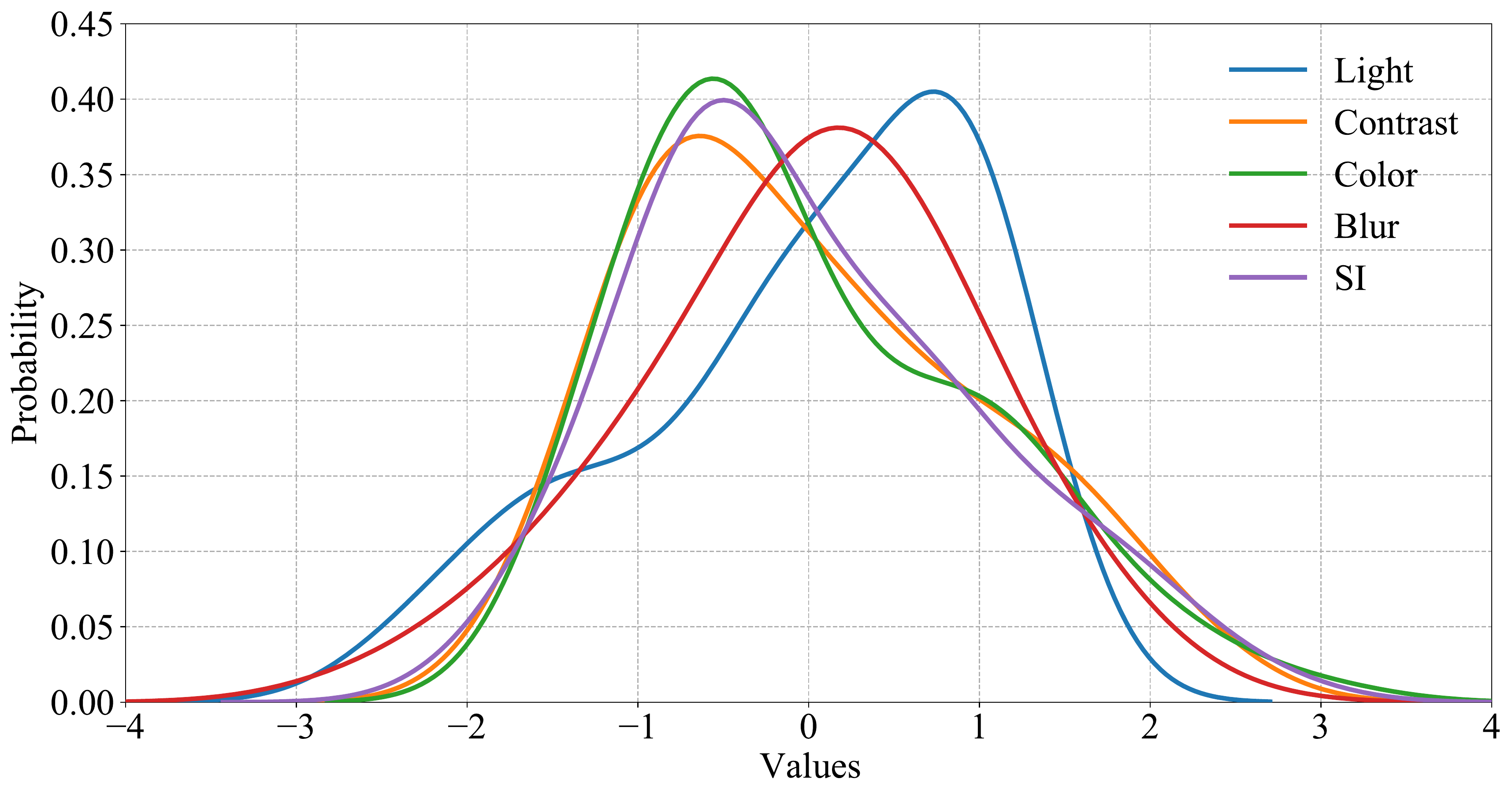}}
    \caption{The normalized probability distributions of 5 quality-related attributes for NSIs and PCIs. The quality-related attributes include light, contrast, colorfulness, blur, and spatial information (SI), the details description of which can be referred to in \cite{hosu2017konstanz}. The distributions are obtained from 10,073 NSIs in the KonIQ-10k IQA database \cite{koniq10k} and (374+740)$\times$120=134,160 PCIs rendered from the SJTU-PCQA database \cite{sjtu-pcqa} and the WPC database \cite{liu2021perceptual} respectively. The 'color' indicates the colorfulness of the images and the 'SI' stands for the texture diversity of the images.}
    \label{fig:diff}
    \vspace{-0.6cm}
\end{figure}


 
\subsection{Performance Discussion}
The final experimental results on the SJTU-PCQA, WPC, and LSPCQA-I databases are shown in Table \ref{tab:experiment}, from which we can make several observations. 1) The proposed method outperforms all NR methods on the all 3 databases. which verifies the effectiveness of the proposed method for predicting the visual quality levels of point clouds. 2) The model-based FR methods achieve relatively better performance, which is because these methods take advantage of reference information and extract features directly from the point cloud.  3) The handcrafted-based NR-VQA methods seem to be less competent for measuring the quality of point clouds. We attempt to give the reasons for such phenomenon. Most handcrafted-based NR-VQA methods are specially developed to tackle the quality assessment for natural scene content. For example, BRISQUE and NIQE operate by measuring the deviations from statistical regularities observed in natural images, however, the statistical regularities do not hold for the point cloud rendering images (PCIs). As seen from Fig. \ref{fig:diff}, PCIs usually contain more geometric shapes, simpler texture, and less colors, which differ greatly from the NSIs in statistics. Therefore, such handcrafted-based NR methods are not suitable for extracting quality-aware information from point cloud projections, which results in lower performance. 4)  Specifically, StairVQA, SimpVQA, and the proposed method are the top 3 among the NR methods on the WPC and LSPCQA-I databases in terms of SRCC. This is because these methods employ powerful DNN models to learn the quality representation from \textbf{both the static and dynamic views, which validates our motivation that the perceived visual quality of point clouds can be better modeled via moving camera videos.}  5) All methods experience a clear performance drop on the WPC and LSPCQA-I databases compared with the SJTU-PCQA database. It can be explained that the WPC and LSPCQA-I databases are more diverse in terms of contents and distortion types. What's more, the distortion levels in the SJTU-PCQA database are relatively coarse-grained, thus making it easier for the PCQA models to distinguish quality differences.  However, the proposed method yields the best performance on the more challenging WPC and LSPCQA-I databases, which further demonstrates the effectiveness of the proposed method.

\begin{figure*}[!tp]
    \centering
    \subfigure[SJTU-PCQA]{
    \begin{minipage}[t]{0.32\linewidth}
    \centering
    \includegraphics[width = 5.7cm]{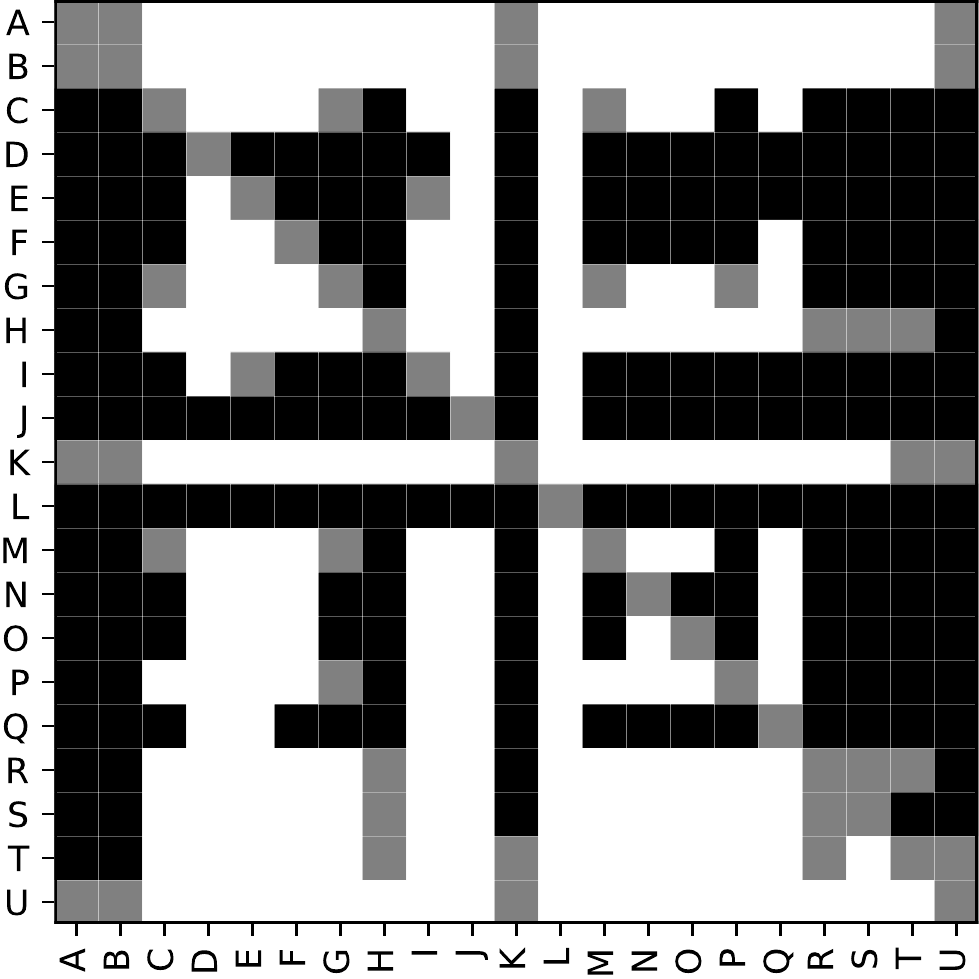}
    \end{minipage}%
    }%
    \subfigure[WPC]{
    \begin{minipage}[t]{0.32\linewidth}
    \centering
    \includegraphics[width = 5.7cm]{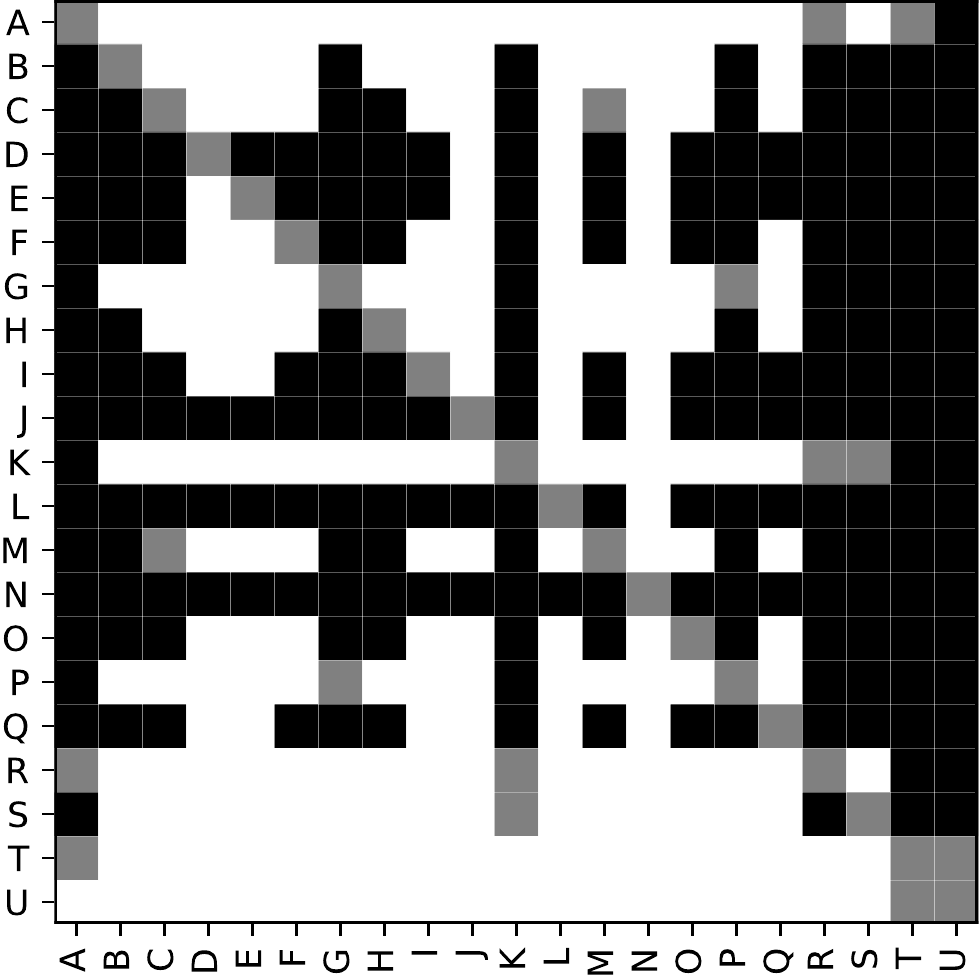}
    \end{minipage}%
    }%
    \subfigure[LSPCQA-I]{
    \begin{minipage}[t]{0.31\linewidth}
    \centering
    \includegraphics[width = 5.7cm]{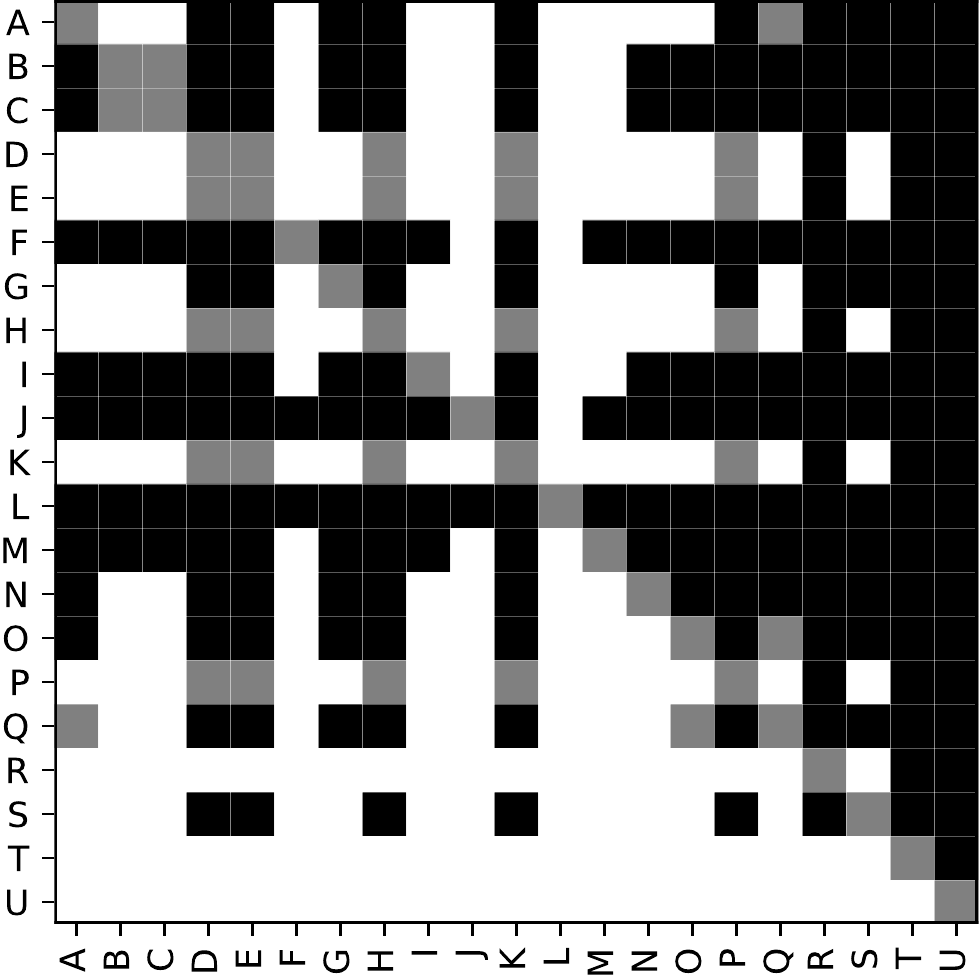}
    \end{minipage}%
    }%
    \caption{{Statistical test results of the proposed method and compared metrics on the SJTU-PCQA, WPC, and LSPCQA-I databases. A black/white block means the row method is statistically worse/better than the column one. A gray block means the row method and the column method are statistically indistinguishable. The metrics are denoted by the same index as in Table \ref{tab:experiment}.}}
    \label{fig:statistic}
    \vspace{-0.5cm}
\end{figure*}

\begin{table*}[!t]
\setlength\tabcolsep{8.5pt}
\caption{{Performance results on over distortion types on the SJTU-PCQA database, where OT represents octree-based compression, CN represents color noise, DS represents down-sampling, DS+CN represents down-sampling and color noise, DS+GGN represents down-sampling and geometry Gaussian noise, GGN represents geometry Gaussian noise, and CN+GGN represents color noise and geometry Gaussian noise.}}
\centering
\begin{tabular}{l|c|c|c|c|c|c|c|c|c|c|c|c|c|c}
\toprule
Distortion    & \multicolumn{2}{c|}{OT}        & \multicolumn{2}{c|}{CN}        & \multicolumn{2}{c|}{DS}        & \multicolumn{2}{c|}{DS+CN}     & \multicolumn{2}{c|}{DS+GGN}    & \multicolumn{2}{c|}{GGN}       & \multicolumn{2}{c}{CN+GGN}   \\ \hline
Method        & SR          & PL          & SR         & PL          & SR          & PL          & SR          & PL          & SR         & PL         & SR         & PL          & SR          & PL         \\ \hline
PCQM          & 0.80          & 0.84          & 0.86          & 0.85          & 0.93          & 0.96          & \textbf{0.97} & {0.94}          & {0.96} & 0.90          & \textbf{0.98} & 0.93          & \textbf{0.99} & 0.93           \\
GraphSIM      & \textbf{0.82} & 0.81          & 0.82          & {0.90} & \textbf{0.96} & \textbf{0.97} & 0.91          & \textbf{0.95} & 0.95          & 0.95          & 0.96          & \textbf{0.97} & 0.97          & \textbf{0.98}   \\
PointSSIM     & 0.80          & \textbf{0.88} & {0.87} & {0.90} & 0.93 & 0.93 & \textbf{0.97} & 0.93          & {0.96} & \textbf{0.97} & \textbf{0.98} & \textbf{0.97} & \textbf{0.99} & 0.96         \\
PQA-net       & 0.81          & 0.82          & 0.84          & 0.83          & 0.91          & 0.92          & 0.93          & 0.91          & 0.89          & 0.89          & 0.95          & {0.96}          & 0.97          & 0.96          \\
3D-NSS        & 0.60          & 0.67          & 0.85          & 0.79          & 0.80          & 0.84          & 0.94          & 0.93          & 0.90          & 0.90          & 0.96          & 0.93          & {0.98}          & 0.94    \\
Proposed & 0.81 & 0.82 & \textbf{0.91} & \textbf{0.92} & {0.95}   & {0.94}          & 0.92 & {0.92} & \textbf{0.97} & {0.94} & 0.91          & 0.91          & {0.96} & {0.96} \\ \bottomrule
\end{tabular}

\label{tab:sjtu}
\vspace{-0.3cm}
\end{table*}

\begin{table}[!t]
\centering
\setlength\tabcolsep{2.4pt}
\caption{{Performance results on over distortion types on the WPC database, where DS represents down-sampling, GN represents geometry and color Gaussian noise, G-PCC represents geometry-based compression, and V-PCC represents video-based compression.} }
\begin{tabular}{l|c|c|c|c|c|c|c|c}
\toprule
Distortion    & \multicolumn{2}{c|}{DS}        & \multicolumn{2}{c|}{GN}        & \multicolumn{2}{c|}{G-PCC}     & \multicolumn{2}{c}{V-PCC}    \\ \hline
Method        & SRCC          & PLCC          & SRCC          & PLCC          & SRCC          & PLCC          & SRCC          & PLCC               \\ \hline
PCQM          & {0.66}          & {0.64}          & {0.89}          & {0.89}          & 0.86          & 0.77          & {0.83}          & {0.78}         \\     
GraphSIM      & 0.56          & 0.57          & 0.79          & 0.81          & 0.75          & 0.74          & 0.71          & 0.70           \\
PointSSIM     & 0.35          & 0.34          & 0.83          & 0.87          & \bf{0.91} & \bf{0.95} & 0.51          & 0.41        \\


PQA-net       & 0.61          & 0.63          & 0.77          & 0.78          & {0.87}          & 0.88          & 0.76          & 0.77     \\
3D-NSS        & 0.55          & 0.51          & 0.81          & 0.83          & 0.86          & 0.87          & 0.49          & 0.47        \\
Proposed & \bf{0.71} & \bf{0.72} & \bf{0.87} & \bf{0.88} & {0.87}          & {0.92}          & \bf{0.89} & \bf{0.88}\\ \bottomrule
\end{tabular}
\label{tab:wpc}
\vspace{-0.3cm}
\end{table}

\begin{table*}[!th]
    \centering
    \renewcommand\tabcolsep{8pt}
    \caption{{Performance results of the cross-database evaluation, where `A' $\rightarrow$ `B' indicates training on the `A' database and testing on the `B' database. ``LP'' indicates the LSPCQA-I database. The best performance is marked in bold.}}
    \begin{tabular}{c|cc|cc|cc|cc|cc|cc}
    \toprule
         \multirow{2}{*}{Methods} & \multicolumn{2}{c|}{SJTU $\rightarrow$ WPC} & \multicolumn{2}{c|}{SJTU $\rightarrow$ LP} & \multicolumn{2}{c|}{WPC $\rightarrow$ SJTU} & \multicolumn{2}{c|}{WPC $\rightarrow$ LP} & \multicolumn{2}{c|}{LP $\rightarrow$ SJTU} & \multicolumn{2}{c}{LP $\rightarrow$ WPC}\\ \cline{2-13}
         & SRCC & PLCC & SRCC & PLCC & SRCC & PLCC & SRCC & PLCC & SRCC & PLCC & SRCC & PLCC \\ \hline
         3D-NSS & 0.1214 & 0.1313 & 0.1071 & 0.1201 & 0.2117 & 0.2034 & 0.1334 & 0.1528 & 0.1001 & 0.0912 & 0.1551 & 0.1466\\
         ResSCNN & 0.2301 & 0.2293 & 0.2233 & 0.2402 & 0.5012 & 0.4954 & 0.3470 & 0.3551 & 0.6009 & 0.5995 & \textbf{0.5108} & \textbf{0.4883} \\
         PQA-net & 0.2211 & 0.2334 & \textbf{0.2677} & \textbf{0.2823} & 0.5411 & 0.6102 & 0.2811  & 0.2969 & 0.5919 & 0.6001 & 0.4772 & 0.4839 \\
         Proposed & \textbf{0.2733} & \textbf{0.3067} & 0.2581 & 0.2742 & \textbf{0.5866} & \textbf{0.6525} & \textbf{0.4103} & \textbf{0.4133} & \textbf{0.6749} & \textbf{0.6791} & 0.4697 & 0.4692 \\
    \bottomrule
    \end{tabular}
    \label{tab:cross}
    \vspace{-0.3cm}
\end{table*}

\subsection{Statistical Test}
{To further analyze the performance of the proposed method, we conduct the statistical test in this section. We follow the same experimental setup as in \cite{statistic-test} and compare the difference between the predicted quality scores with the subjective ratings. All possible pairs of models are tested and the results are listed in Fig. \ref{fig:statistic}. The method index is of the same order as in Table. \ref{tab:experiment}. 
It can be observed that the proposed method is significantly superior to most comparing methods. More specifically, the proposed method significantly outperforms 16, 19, and 20 compared methods on the SJTU-PCQA, WPC, and LSPCQA-I databases respectively.}

\subsection{Distortion-specific Test}
{
We further conduct the distortion-specific experiment. The experimental performance is exhibited in Table \ref{tab:sjtu} and Table \ref{tab:wpc}.
Several findings can be obtained: 1) All the PCQA models demonstrate lower performance when dealing with octree-based compression on the SJTU-PCQA database, which suggests that determining the quality levels of octree-based compression presents significant challenges. 2) On the WPC database, most PCQA methods appear sensitive to G-PCC compression but less so to V-PCC compression except PQA-net and the proposed method. This is because V-PCC leverages commonly used video-based coding standards like H.264 for compression. Conversely, most of the compared PCQA methods (excluding PQA-net and the proposed method) derive features straight from points instead of projections, thereby exhibiting superior performance on G-PCC over V-PCC. 3) When observing the performance on the WPC database, it's clear that our proposed method outperforms others in evaluating 3 out of the 4 distortion types in terms of SRCC, which reaffirms the proposed method's effectiveness and robustness.}

\subsection{Cross-database Validation}
{The experimental results for cross-database validation are presented in Table \ref{tab:cross}, from which several key observations can be deduced. a) Evidently, PCQA models that are trained on databases of larger scale tend to exhibit superior performance compared to those trained on smaller databases. To illustrate, all LP 
$\rightarrow$ WPC PCQA models markedly surpass the performance of SJTU $\rightarrow$ WPC PCQA models. This underscores the advantage of having a more extensive set of training samples in enhancing the efficacy of PCQA models. b) In a holistic view, the method proposed in this study consistently demonstrates a performance that is relatively more commendable than other PCQA models, reaffirming the robust generalization capability of the introduced approach.}

\subsection{Ablation Studies}
\subsubsection{Spatial and Temporal Contributions}
Since the proposed method incorporates both temporal and spatial information for quality prediction, we try to quantify the contributions of the temporal and spatial components. 
The default experimental setup is maintained and the experimental results are shown in Table \ref{tab:ablation}. The separate ResNet50 and SlowFast models are inferior to the proposed model, which implies that both spatial and temporal information make contributions. With closer inspections, the spatial information makes more contributions to the final results.

\begin{table}[!tp]
\caption{Performance results of the spatial and temporal contributions on the SJTU-PCQA and WPC databases, where `S' represents spatial features and 'T' represents temporal features. Best in bold.}
\renewcommand\tabcolsep{12pt}
    \centering
    \begin{tabular}{c|cc|cc}
    \toprule
    \multirow{2}{*}{Model} & \multicolumn{2}{c|}{SJTU-PCQA} & \multicolumn{2}{c}{WPC} \\ \cline{2-5}
                & SRCC$\uparrow$    & PLCC $\uparrow$  & SRCC $\uparrow$   & PLCC$\uparrow$ \\ \hline
        S   &0.8188 &0.8452 & 0.7906 & 0.7851\\
        T  &0.3443 &0.5985 & 0.1815 & 0.2509\\ 
        S+T &\bf{0.8611} &\bf{0.8702}& \bf{0.8012} & \bf{0.8001}\\
    \bottomrule
    \end{tabular}
    \vspace{-0.25cm}
    \label{tab:ablation}
\end{table}

\begin{table}[!tp]
\caption{Performance comparison of employing different backbones on the SJTU-PCQA and WPC databases, where `MNetV2' stands for MobileNetV2. The backbones marked with gray color are the proposed selection.  Best in bold.}
\renewcommand\tabcolsep{3pt}
    \centering
    \begin{tabular}{c|c|c|cc|cc}
    \toprule
    \multirow{2}{*}{Para. (M)} & \multicolumn{2}{c|}{Backbone}&  \multicolumn{2}{c|}{SJTU-PCQA} & \multicolumn{2}{c}{WPC} \\ \cline{2-7}
        &  Spatial & Temporal       & SRCC$\uparrow$    & PLCC $\uparrow$  &  SRCC $\uparrow$   & PLCC$\uparrow$ \\ \hline
    6.25   & MNetV2 & X3D  &0.8066 &0.8121 & 0.7333 & 0.7422 \\ 
    26.78    &ResNet50 & X3D     &0.8514 &{0.8758} & 0.7822 & 0.7920 \\
    36.51   & MNetV2 & SlowFast  &0.8204 &0.8405 & 0.7809 & 0.7701 \\ 
    55.44    & ResNet34 & SlowFast &0.8313 
        &\bf{0.8877} & 0.7856 & 0.7822\\ 
        \rowcolor{mygray}
     58.36   &  ResNet50 & SlowFast &\bf{0.8611} &0.8702& \bf{0.8012} & \bf{0.8001}\\ 
    78.21    & ResNet101 & SlowFast &0.8211 &0.8244& 0.7764 & 0.7551\\
    171.99    & VGG16 & SlowFast    &0.8337 &0.8472& 0.7885 & 0.7986\\
    \bottomrule
    \end{tabular}
    \label{tab:backbone}
\end{table}

\subsubsection{Backbones Comparison}
In addition, we replace the 2D-CNN and 3D-CNN backbones with other mainstream models for validation. Specifically, we replace the ResNet50 backbone with ResNet34, ResNet101 \cite{he2016deep}, VGG16 \cite{simonyan2014very}, and MobileNetV2 \cite{sandler2018mobilenetv2} as spatial feature extractor. We replace the pre-trained SlowFast model \cite{feichtenhofer2019slowfast} with the lightweight pre-trained X3D \cite{feichtenhofer2020x3d} as the temporal feature extractor. The experimental results and the consuming parameters are exhibited in Table \ref{tab:backbone}. First, by comparing the performance of different backbones, the proposed selection is optimal for the PCQA tasks on the validated databases. Second, although models employing other mainstream backbones achieve lower performance than the proposed method, they are still competitive among the previously developed PCQA methods, which further verifies the effectiveness of the proposed structure. The MobileNetV2 and X3D combination model only takes up 6.25 million parameters and maintains relatively good performance, which demonstrate the potential of applying the proposed model to the lightweight platforms.

\begin{table}[!t]
\caption{{ Performance results of different angular separation. Best in bold.}}
\renewcommand\tabcolsep{10pt}
    \centering
    \begin{tabular}{c|cc|cc}
    \toprule
    Angular & \multicolumn{2}{c|}{SJTU-PCQA} & \multicolumn{2}{c}{WPC} \\ \cline{2-5}
      Separation  & SRCC$\uparrow$    & PLCC $\uparrow$  & SRCC $\uparrow$   & PLCC$\uparrow$ \\ \hline
        12º   & \bf{0.8509} & \bf{0.8635} & \bf{0.7968} & \bf{0.7976} \\
        18º   & 0.8421 & 0.8534 & 0.7890 & 0.7898 \\
        24º   & 0.8365 & 0.8478 & 0.7924 & 0.7831 \\
        30º   & 0.8302 & 0.8416 & 0.7767 & 0.7775 \\
        36º   & 0.8340 & 0.8455 & 0.7709 & 0.7716 \\
        45º   & 0.8189 & 0.8304 & 0.7653 & 0.7661 \\
        72º   & 0.8194 & 0.8428 & 0.7701 & 0.7720 \\
    \bottomrule
    \end{tabular}
    \vspace{-0.25cm}
    \label{tab:angular_separation}
\end{table}

\begin{figure}[t]
    \centering
    \includegraphics[width=.9\linewidth]{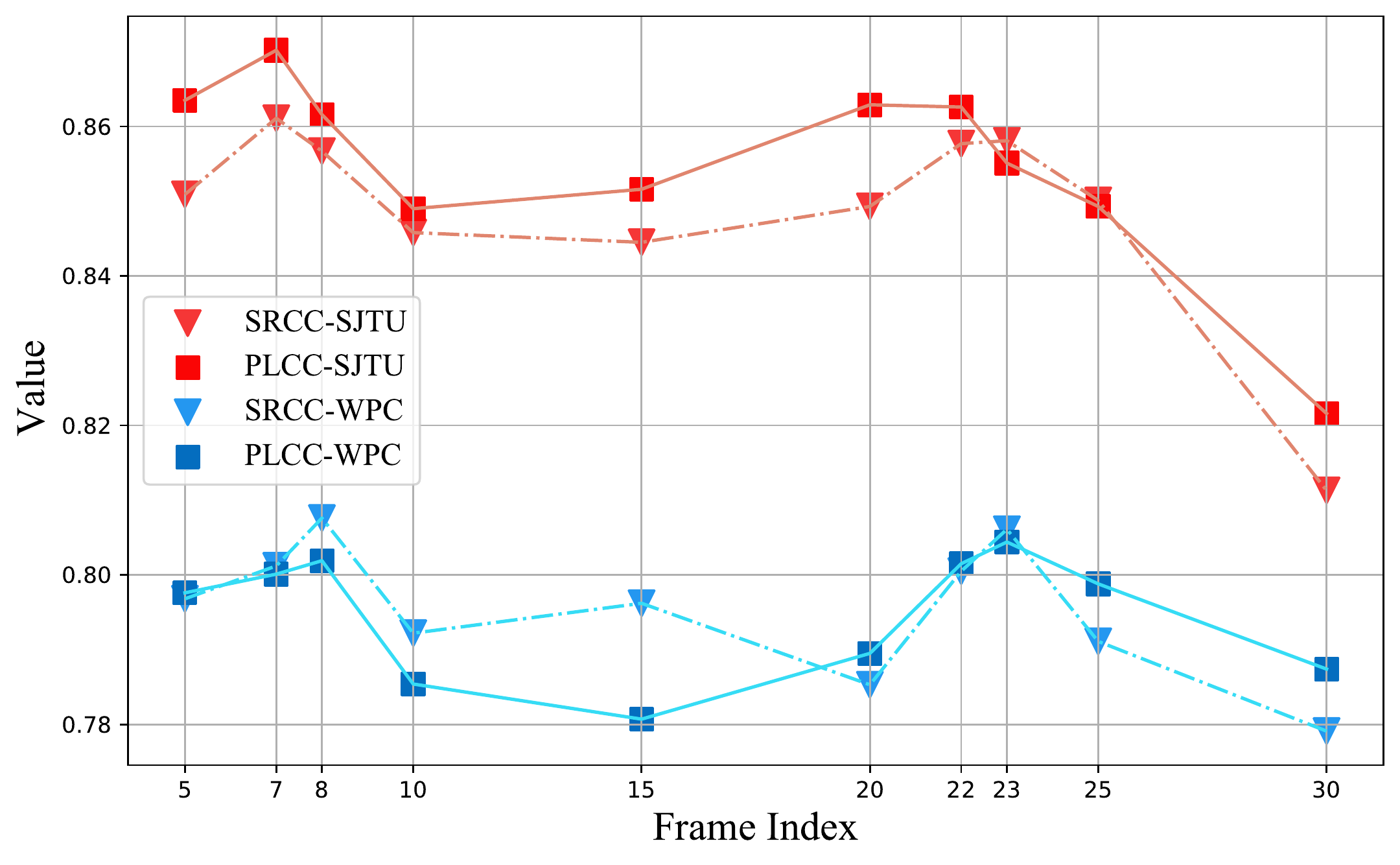}
    \caption{Performance comparison of different key frame selection on the SJTU-PCQA and WPC databases. The 7-th frame is the proposed key frame selection.}
    \label{fig:keyframe}
\end{figure}

\subsubsection{Angular Separations Comparison}
{In this section, we explore the efficacy of employing larger angular separations. A more extensive angular separation implies a larger frame gap, potentially affecting the temporal feature extraction. We've selected angular separations of 18º, 24º, 30º, 36º, 45º, and 72º for comparison. These separations correspond to clips with 20 FPS, 15 FPS, 10 FPS, 8 FPS, and 5 FPS, respectively, for feature extraction. The results of these experiments can be found in Table \ref{tab:angular_separation}. A decline in performance with increasing angular separation is evident from the table. The broader angular separations might increase the disparity between consecutive frames, thus negatively influencing the capture of continuous temporal quality features.}

\subsubsection{Key Frame Selection}
{ In this section, we try to find out the influence of the viewpoint-max-distance strategy. We compare the performance of the 7-th/8-th frames (midpoint of 1-15) and the 22-nd/23-rd frames (midpoint of 15-30) of each clip, which has the most distant viewpoints in the proposed projection setting. Besides, we manually choose the 5-th, 10-th, 15-th, 20-th, 25-th, and 30-th frames as baselines. The experimental results are shown in Fig. \ref{fig:keyframe}, from which we can observe that choosing the 30-th frame of each clip as the key frame causes a clear performance drop. It is because the camera returns to the same starting point after each clip and the 30-th frame is captured from close viewpoints, thus covering less content than other selection. Moreover, the VMD-sampled 7-th/8-th/22-nd/23-rd key frames selections can help improve the performance since these viewpoints can cover more point cloud content.}

\begin{table}[t]
\caption{Performance results of pathway comparison on the SJTU-PCQA and WPC databases. The mark \checkmark in the pathways column indicates that the corresponding pathway is used. Best in bold.}
\renewcommand\tabcolsep{4.8pt}
    \centering
    \begin{tabular}{c|cccc|cc|cc}
    \toprule
    \multirow{2}{*}{Model}&\multicolumn{4}{c|}{Pathway} & \multicolumn{2}{c|}{SJTU-PCQA} & \multicolumn{2}{c}{WPC} \\ \cline{2-9}
    & $\theta_{A}$ & $\theta_{B}$ & $\theta_{C}$ & $\theta_{D}$ & SRCC$\uparrow$ & PLCC $\uparrow$& SRCC$\uparrow$ & PLCC$\uparrow$ \\ \hline
    $P1$ & \checkmark & $\times$ & $\times$ & $\times$ & 0.7921 & 0.8320 & 0.7335 & 0.7299 \\
    $P2$ & \checkmark & \checkmark & $\times$ & $\times$ & 0.8212 & 0.8422 & 0.7489 & 0.7336 \\
    $P3$ & \checkmark & \checkmark & \checkmark & $\times$ & 0.8304 & 0.8471 & 0.7739 & 0.7785 \\
    $P4$ & \checkmark & \checkmark & $\times$ & \checkmark & 0.8433 & 0.8630 & 0.7866 & 0.7741 \\
    $P5$ & \checkmark & $\times$ & \checkmark & \checkmark & 0.8480 & \bf{0.8702} & 0.7591 & 0.7705 \\
    $P6$ & $\times$ & \checkmark & \checkmark & \checkmark & 0.8380 & 0.8650 & 0.7642 & 0.7700 \\
    $P7$& \checkmark &\checkmark &\checkmark &\checkmark & \bf{0.8611} & \bf{0.8702} &\bf{0.8012} &\bf{0.8001}\\
    \bottomrule
    \end{tabular}
    \vspace{-0.3cm}
    \label{tab:pathway}
\end{table}

\subsubsection{Pathway Comparison}
The proposed method employs four pathways to capture videos. Although the pathways are quite different, the capture clips may still share similar visual contents. It is necessary to investigate whether there is redundancy among the pathways and quantify the contributions of each pathway. Therefore, in this section, we use 3, 2, and 1 pathways for the experiment and list the performance results in Table \ref{tab:pathway}. First, by comparing the performance of model $P1$, $P2$, $P3$, and $P7$, it can be viewed that the experimental performance is consistently improved with the increasing number of pathways, which indicates that capturing more video clips is beneficial for enhancing the quality understanding of the model. Second, the $P3$ $\sim$ $P6$ models are all inferior to the $P7$ model, which implies that all four pathways make contributions to the final results. Moreover, the P6 model obtains the lowest performance among the $P3$ $\sim$ $P6$ models, which suggests that pathway $\theta_{A}$ devotes more among the four pathways.

\subsection{Computational Efficiency}
\begin{figure}[!t]
    \centering
    \includegraphics[width=.9\linewidth]{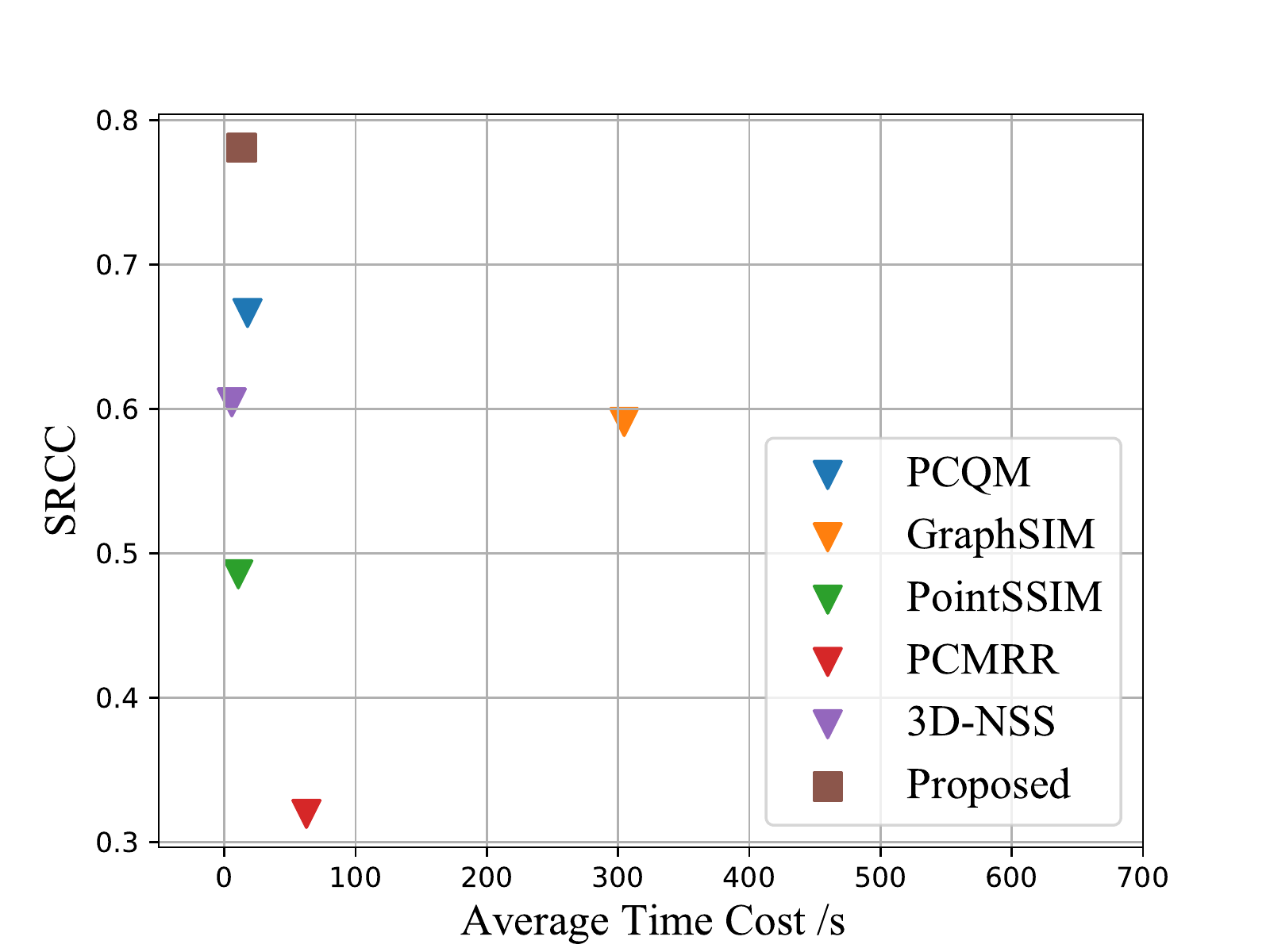}
    \caption{The results of computational efficiency comparison. The SRCC indicates the average SRCC values of corresponding PCQA methods on the SJTU-PCQA, WPC, and LSPCQA-I databases. The average time cost represents the average running time of the PCQA methods to evaluate a single point cloud.}
    \label{fig:computational}
    \vspace{-0.3cm}
\end{figure}

Since the proposed method needs to first capture the videos and then predict the visual quality of point clouds from the video, we are also interested in the computational efficiency of the proposed method. Considering that the projection-based methods share the time consumption of generating videos, we mainly select the model-based methods for comparison, which include PCQM, GraphSIM, PointSSIM, PCMRR, and 3D-NSS. All the algorithms are implemented on a laptop with AMD Ryzen 7 5800H @ 3.20 GHz CPU, 16G RAM, and NVIDIA Geforce RTX 3050Ti GPU on the Windows platform. We report the average running time per point cloud on the SJTU-PCQA, WPC, and LSPCQA-I databases for each PCQA method. It's worth mentioning that the running time for the proposed method includes the video capture time consumption and inference time consumption. The computational efficiency comparison results are illustrated in Fig. \ref{fig:computational} and the detailed time cost is listed in Table \ref{tab:computational}. Although PointSSIM and 3D-NSS take less time than the proposed method, their performance is considerably lower than the proposed method. The proposed method achieves first place on average SRCC values while maintaining competitive computational efficiency, which implies the proposed method is capable of handling practical applications.

\begin{table}[!t]
\caption{The detailed time cost of comparing PCQA methods.}
\renewcommand\arraystretch{1.2}
\renewcommand\tabcolsep{3pt}
    \centering
    \centering
    \begin{tabular}{c|cccccc} \toprule
         Method & PCQM & GraphSIM & PointSSIM & PCMRR & 3D-NSS & Proposed \\ \hline
         Time/s & 17.63 & 304.64 & 10.63 & 62.53 & 5.68 & 13.15 \\
         \bottomrule
    \end{tabular}
    \label{tab:computational}
    \vspace{-0.3cm}
\end{table}

\section{Conclusion}
\label{sec:conclusion}
This paper proposes a no-reference quality assessment method for point cloud by evaluating point cloud from moving camera videos. We hold the assumption that people tend to perceive the point cloud through both static and dynamic views. However, previous projection-based PCQA methods only evaluate the rendered 2D projections via fixed viewpoints. Therefore, the proposed method first generates the videos by rotating the camera along four designed pathways around the point clouds. Then the proposed method extracts spatial and temporal information from the key frames and video clips by employing trainable ResNet50 and pre-trained SlowFast R50 model. In this way, the proposed method is able to incorporate the quality-aware information from both static and dynamic views and gains an advantage over the previous projection-based PCQA methods. The experimental results show that the proposed method achieves significantly better performance than all NR-PCQA methods and the performance is even indistinguishable from FR-PCQA methods, which shows the effectiveness of the proposed method. Furthermore, the ablation study verifies the rationality of the proposed structure, and the detailed analysis is given as well.

 




\bibliographystyle{IEEEtran}
\bibliography{output}

\begin{thebibliography}{10}
\providecommand{\url}[1]{#1}
\csname url@samestyle\endcsname
\providecommand{\newblock}{\relax}
\providecommand{\bibinfo}[2]{#2}
\providecommand{\BIBentrySTDinterwordspacing}{\spaceskip=0pt\relax}
\providecommand{\BIBentryALTinterwordstretchfactor}{4}
\providecommand{\BIBentryALTinterwordspacing}{\spaceskip=\fontdimen2\font plus
\BIBentryALTinterwordstretchfactor\fontdimen3\font minus \fontdimen4\font\relax}
\providecommand{\BIBforeignlanguage}[2]{{%
\expandafter\ifx\csname l@#1\endcsname\relax
\typeout{** WARNING: IEEEtran.bst: No hyphenation pattern has been}%
\typeout{** loaded for the language `#1'. Using the pattern for}%
\typeout{** the default language instead.}%
\else
\language=\csname l@#1\endcsname
\fi
#2}}
\providecommand{\BIBdecl}{\relax}
\BIBdecl

\bibitem{xiong2021augmented}
J.~Xiong, E.-L. Hsiang, Z.~He, T.~Zhan, and S.-T. Wu, ``Augmented reality and virtual reality displays: emerging technologies and future perspectives,'' \emph{Light: Science \& Applications}, vol.~10, no.~1, pp. 1--30, 2021.

\bibitem{kang2020review}
Z.~Kang, J.~Yang, Z.~Yang, and S.~Cheng, ``A review of techniques for 3d reconstruction of indoor environments,'' \emph{ISPRS International Journal of Geo-Information}, vol.~9, no.~5, p. 330, 2020.

\bibitem{ning2021survey}
H.~Ning, H.~Wang, Y.~Lin, W.~Wang, S.~Dhelim, F.~Farha, J.~Ding, and M.~Daneshmand, ``A survey on metaverse: the state-of-the-art, technologies, applications, and challenges,'' \emph{arXiv preprint arXiv:2111.09673}, 2021.

\bibitem{chen2013point}
J.-Y. Chen, C.-H. Lin, P.-C. Hsu, and C.-H. Chen, ``Point cloud encoding for 3d building model retrieval,'' \emph{IEEE transactions on multimedia}, vol.~16, no.~2, pp. 337--345, 2013.

\bibitem{sim2008compression}
J.-Y. Sim and S.-U. Lee, ``Compression of 3-d point visual data using vector quantization and rate-distortion optimization,'' \emph{IEEE Transactions on Multimedia}, vol.~10, no.~3, pp. 305--315, 2008.

\bibitem{park2008multiscale}
S.-B. Park and S.-U. Lee, ``Multiscale representation and compression of 3-d point data,'' \emph{IEEE Transactions on Multimedia}, vol.~11, no.~1, pp. 177--183, 2008.

\bibitem{javaheri2020point}
A.~Javaheri, C.~Brites, F.~Pereira, and J.~Ascenso, ``Point cloud rendering after coding: Impacts on subjective and objective quality,'' \emph{IEEE Transactions on Multimedia}, vol.~23, pp. 4049--4064, 2020.

\bibitem{de2018graph}
P.~de~Oliveira~Rente, C.~Brites, J.~Ascenso, and F.~Pereira, ``Graph-based static 3d point clouds geometry coding,'' \emph{IEEE Transactions on Multimedia}, vol.~21, no.~2, pp. 284--299, 2018.

\bibitem{p2point}
P.~Cignoni, C.~Rocchini, and R.~Scopigno, ``Metro: Measuring error on simplified surfaces,'' \emph{Computer Graphics Forum}, vol.~17, no.~2, pp. 167--174, 1998.

\bibitem{p2plane}
R.~Mekuria and P.~Cesar, ``Mp3dg-pcc, open source software framework for implementation and evaluation of point cloud compression.''\hskip 1em plus 0.5em minus 0.4em\relax Association for Computing Machinery, 2016, p. 1222–1226.

\bibitem{m1}
G.~Lavoué, ``A multiscale metric for 3d mesh visual quality assessment,'' \emph{Computer Graphics Forum}, vol.~30, no.~5, pp. 1427--1437, 2011.

\bibitem{ff2_roughness}
K.~Wang, F.~Torkhani, and A.~Montanvert, ``A fast roughness-based approach to the assessment of 3d mesh visual quality,'' \emph{Computers and Graphics}, vol.~36, no.~7, pp. 808--818, 2012.

\bibitem{p2mesh}
D.~Tian, H.~Ochimizu, C.~Feng, R.~Cohen, and A.~Vetro, ``Geometric distortion metrics for point cloud compression,'' in \emph{IEEE International Conference on Image Processing}, 2017, pp. 3460--3464.

\bibitem{angular}
E.~Alexiou and T.~Ebrahimi, ``Point cloud quality assessment metric based on angular similarity,'' in \emph{IEEE International Conference on Multimedia and Expo}, 2018, pp. 1--6.

\bibitem{pcqa2}
A.~Javaheri, C.~Brites, F.~Pereira, and J.~Ascenso, ``A generalized hausdorff distance based quality metric for point cloud geometry,'' in \emph{International Conference on Quality of Multimedia Experience}, 2020, pp. 1--6.

\bibitem{dame}
L.~Váša and J.~Rus, ``Dihedral angle mesh error: a fast perception correlated distortion measure for fixed connectivity triangle meshes,'' \emph{Computer Graphics Forum}, vol.~31, no.~5, pp. 1715--1724, 2012.

\bibitem{pcqa3}
E.~Alexiou and T.~Ebrahimi, ``Towards a point cloud structural similarity metric,'' in \emph{IEEE International Conference on Multimedia Expo Workshops}, 2020, pp. 1--6.

\bibitem{tian-color}
D.~Tian and G.~AlRegib, ``Batex3: Bit allocation for progressive transmission of textured 3-d models,'' \emph{IEEE Transactions on Circuits and Systems for Video Technology}, vol.~18, no.~1, pp. 23--35, 2008.

\bibitem{guo-color}
J.~Guo, V.~Vidal, I.~Cheng, A.~Basu, A.~Baskurt, and G.~Lavoue, ``Subjective and objective visual quality assessment of textured 3d meshes,'' \emph{ACM Transactions on Applied Perception}, vol.~14, no.~2, 2016.

\bibitem{pcqm}
G.~Meynet, Y.~Nehmé, J.~Digne, and G.~Lavoué, ``Pcqm: A full-reference quality metric for colored 3d point clouds,'' in \emph{International Conference on Quality of Multimedia Experience}, 2020, pp. 1--6.

\bibitem{liu2021reduced}
Q.~Liu, H.~Yuan, R.~Hamzaoui, H.~Su, J.~Hou, and H.~Yang, ``Reduced reference perceptual quality model with application to rate control for video-based point cloud compression,'' \emph{IEEE Transactions on Image Processing}, vol.~30, pp. 6623--6636, 2021.

\bibitem{zhang2022no}
Z.~Zhang, W.~Sun, X.~Min, T.~Wang, W.~Lu, and G.~Zhai, ``No-reference quality assessment for 3d colored point cloud and mesh models,'' \emph{IEEE Transactions on Circuits and Systems for Video Technology}, 2022.

\bibitem{zhang2022mm}
Z.~Zhang, W.~Sun, X.~Min, Q.~Zhou, J.~He, Q.~Wang, and G.~Zhai, ``Mm-pcqa: Multi-modal learning for no-reference point cloud quality assessment,'' \emph{International Joint Conferences on Artificial Intelligence}, 2023.

\bibitem{zhang2023eep}
Z.~Zhang, W.~Sun, Y.~Zhou, W.~Lu, Y.~Zhu, X.~Min, and G.~Zhai, ``Eep-3dqa: Efficient and effective projection-based 3d model quality assessment,'' \emph{IEEE International Conference on Multimedia and Expo}, 2023.

\bibitem{zhang2023gms}
Z.~Zhang, W.~Sun, H.~Wu, Y.~Zhou, C.~Li, X.~Min, G.~Zhai, and W.~Lin, ``Gms-3dqa: Projection-based grid mini-patch sampling for 3d model quality assessment,'' \emph{arXiv preprint arXiv:2306.05658}, 2023.

\bibitem{alexiou2020pointssim}
E.~Alexiou and T.~Ebrahimi, ``Towards a point cloud structural similarity metric,'' in \emph{2020 IEEE International Conference on Multimedia \& Expo Workshops}.\hskip 1em plus 0.5em minus 0.4em\relax IEEE, 2020, pp. 1--6.

\bibitem{yang2020graphsim}
Q.~Yang, Z.~Ma, Y.~Xu, Z.~Li, and J.~Sun, ``Inferring point cloud quality via graph similarity,'' \emph{IEEE Transactions on Pattern Analysis and Machine Intelligence}, 2020.

\bibitem{meynet2020pcqm}
G.~Meynet, Y.~Nehm{\'e}, J.~Digne, and G.~Lavou{\'e}, ``Pcqm: A full-reference quality metric for colored 3d point clouds,'' in \emph{2020 Twelfth International Conference on Quality of Multimedia Experience}.\hskip 1em plus 0.5em minus 0.4em\relax IEEE, 2020, pp. 1--6.

\bibitem{zhang2021no}
Z.~Zhang, W.~Sun, X.~Min, T.~Wang, W.~Lu, W.~Zhu, and G.~Zhai, ``A no-reference visual quality metric for 3d color meshes,'' in \emph{2021 IEEE International Conference on Multimedia \& Expo Workshops}.\hskip 1em plus 0.5em minus 0.4em\relax IEEE, 2021, pp. 1--6.

\bibitem{sjtu-pcqa}
Q.~Yang, H.~Chen, Z.~Ma, Y.~Xu, R.~Tang, and J.~Sun, ``Predicting the perceptual quality of point cloud: A {3D}-to-{2D} projection-based exploration,'' \emph{IEEE Transactions on Multimedia}, pp. 1--1, 2020.

\bibitem{pcqa_database2}
E.~M. Torlig, E.~Alexiou, T.~A. Fonseca, R.~L. de~Queiroz, and T.~Ebrahimi, ``{A novel methodology for quality assessment of voxelized point clouds},'' in \emph{Applications of Digital Image Processing XLI}, vol. 10752.\hskip 1em plus 0.5em minus 0.4em\relax International Society for Optics and Photonics, 2018, pp. 174 -- 190.

\bibitem{liu2021pqa}
Q.~Liu, H.~Yuan, H.~Su, H.~Liu, Y.~Wang, H.~Yang, and J.~Hou, ``{PQA-N}et: Deep no reference point cloud quality assessment via multi-view projection,'' \emph{IEEE Transactions on Circuits and Systems for Video Technology}, 2021.

\bibitem{li2022blindly}
B.~Li, W.~Zhang, M.~Tian, G.~Zhai, and X.~Wang, ``Blindly assess quality of in-the-wild videos via quality-aware pre-training and motion perception,'' \emph{IEEE Transactions on Circuits and Systems for Video Technology}, 2022.

\bibitem{sun2022deep}
W.~Sun, X.~Min, W.~Lu, and G.~Zhai, ``A deep learning based no-reference quality assessment model for ugc videos,'' in \emph{Proceedings of the 30th ACM International Conference on Multimedia}, 2022, pp. 856--865.

\bibitem{liu2021perceptual}
Q.~Liu, H.~Su, Z.~Duanmu, W.~Liu, and Z.~Wang, ``Perceptual quality assessment of colored 3d point clouds,'' \emph{IEEE Transactions on Visualization and Computer Graphics}, 2022.

\bibitem{pcqa-large-scale}
Y.~Liu, Q.~Yang, Y.~Xu, and L.~Yang, ``Point cloud quality assessment: Dataset construction and learning-based no-reference metric,'' \emph{ACM Transactions on Multimedia Computing, Communications, and Applications}, 2022.

\bibitem{yang2022mped}
Q.~Yang, Y.~Zhang, S.~Chen, Y.~Xu, J.~Sun, and Z.~Ma, ``Mped: Quantifying point cloud distortion based on multiscale potential energy discrepancy,'' \emph{IEEE Transactions on Pattern Analysis and Machine Intelligence}, vol.~45, no.~5, pp. 6037--6054, 2022.

\bibitem{zhang2022evaluating}
\BIBentryALTinterwordspacing
Y.~Zhang, Q.~Yang, Y.~Zhou, X.~Xu, L.~Yang, and Y.~Xu, ``Evaluating point cloud quality via transformational complexity,'' \emph{arXiv preprint arXiv:2210.04671}, 2022. [Online]. Available: \url{https://arxiv.org/abs/2210.04671}
\BIBentrySTDinterwordspacing

\bibitem{yang2022no}
Q.~Yang, Y.~Liu, S.~Chen, Y.~Xu, and J.~Sun, ``No-reference point cloud quality assessment via domain adaptation,'' in \emph{Proceedings of the IEEE/CVF Conference on Computer Vision and Pattern Recognition}, 2022, pp. 21\,179--21\,188.

\bibitem{tu2022v}
R.~Tu, G.~Jiang, M.~Yu, T.~Luo, Z.~Peng, and F.~Chen, ``V-pcc projection based blind point cloud quality assessment for compression distortion,'' \emph{IEEE Transactions on Emerging Topics in Computational Intelligence}, vol.~7, no.~2, pp. 462--473, 2022.

\bibitem{shan2023gpanet}
Z.~Shan, Q.~Yang, R.~Ye, Y.~Zhang, Y.~Xu, X.~Xu, and S.~Liu, ``{GPA-N}et: No-reference point cloud quality assessment with multi-task graph convolutional network,'' \emph{IEEE Transactions on Visualization and Computer Graphics}, 2023.

\bibitem{vmaf}
\BIBentryALTinterwordspacing
``Vmaf - video multi-method assessment fusion,'' May. 30, 2022. [Online]. Available: \url{https://github.com/Netflix/vmaf}
\BIBentrySTDinterwordspacing

\bibitem{sun2023analysis}
W.~Sun, W.~Wen, X.~Min, L.~Lan, G.~Zhai, and K.~Ma, ``Analysis of video quality datasets via design of minimalistic video quality models,'' \emph{arXiv preprint arXiv:2307.13981}, 2023.

\bibitem{mittal2015completely}
A.~Mittal, M.~A. Saad, and A.~C. Bovik, ``A completely blind video integrity oracle,'' \emph{IEEE Transactions on Image Processing}, vol.~25, no.~1, pp. 289--300, 2015.

\bibitem{saad2014blind}
M.~A. Saad, A.~C. Bovik, and C.~Charrier, ``Blind prediction of natural video quality,'' \emph{IEEE Transactions on Image Processing}, vol.~23, no.~3, pp. 1352--1365, 2014.

\bibitem{korhonen2019two}
J.~Korhonen, ``Two-level approach for no-reference consumer video quality assessment,'' \emph{IEEE Transactions on Image Processing}, vol.~28, no.~12, pp. 5923--5938, 2019.

\bibitem{tu2021ugc}
Z.~Tu, Y.~Wang, N.~Birkbeck, B.~Adsumilli, and A.~C. Bovik, ``Ugc-vqa: Benchmarking blind video quality assessment for user generated content,'' \emph{IEEE Transactions on Image Processing}, vol.~30, pp. 4449--4464, 2021.

\bibitem{li2019quality}
D.~Li, T.~Jiang, and M.~Jiang, ``Quality assessment of in-the-wild videos,'' in \emph{Proceedings of the 27th ACM International Conference on Multimedia}, 2019, pp. 2351--2359.

\bibitem{tu2021rapique}
Z.~Tu, X.~Yu, Y.~Wang, N.~Birkbeck, B.~Adsumilli, and A.~C. Bovik, ``Rapique: Rapid and accurate video quality prediction of user generated content,'' \emph{IEEE Open Journal of Signal Processing}, vol.~2, pp. 425--440, 2021.

\bibitem{sun2021deep}
W.~Sun, T.~Wang, X.~Min, F.~Yi, and G.~Zhai, ``Deep learning based full-reference and no-reference quality assessment models for compressed ugc videos,'' in \emph{2021 IEEE International Conference on Multimedia \& Expo Workshops}.\hskip 1em plus 0.5em minus 0.4em\relax IEEE, 2021, pp. 1--6.

\bibitem{lu2022deep}
W.~Lu, W.~Sun, X.~Min, W.~Zhu, Q.~Zhou, J.~He, Q.~Wang, Z.~Zhang, T.~Wang, and G.~Zhai, ``Deep neural network for blind visual quality assessment of 4k content,'' \emph{IEEE Transactions on Broadcasting}, 2022.

\bibitem{zhang2023md}
Z.~Zhang, W.~Wu, W.~Sun, D.~Tu, W.~Lu, X.~Min, Y.~Chen, and G.~Zhai, ``Md-vqa: Multi-dimensional quality assessment for ugc live videos,'' in \emph{Proceedings of the IEEE/CVF Conference on Computer Vision and Pattern Recognition}, 2023, pp. 1746--1755.

\bibitem{ssim}
Z.~Wang, A.~Bovik, H.~Sheikh, and E.~Simoncelli, ``Image quality assessment: from error visibility to structural similarity,'' \emph{IEEE Transactions on Image Processing}, vol.~13, no.~4, pp. 600--612, 2004.

\bibitem{niqe}
A.~{Mittal}, R.~{Soundararajan}, and A.~C. {Bovik}, ``Making a “completely blind” image quality analyzer,'' \emph{IEEE Signal Processing Letters}, vol.~20, no.~3, pp. 209--212, 2013.

\bibitem{Zhou2018}
Q.-Y. Zhou, J.~Park, and V.~Koltun, ``{Open3D}: {A} modern library for {3D} data processing,'' \emph{arXiv:1801.09847}, 2018.

\bibitem{yang2020inferring}
Q.~Yang, Z.~Ma, Y.~Xu, Z.~Li, and J.~Sun, ``Inferring point cloud quality via graph similarity,'' \emph{IEEE Transactions on Pattern Analysis and Machine Intelligence}, 2020.

\bibitem{viola2020reduced}
I.~Viola and P.~Cesar, ``A reduced reference metric for visual quality evaluation of point cloud contents,'' \emph{IEEE Signal Processing Letters}, vol.~27, pp. 1660--1664, 2020.

\bibitem{brisque}
A.~Mittal, A.~K. Moorthy, and A.~C. Bovik, ``No-reference image quality assessment in the spatial domain,'' \emph{IEEE Transactions on Image Processing}, vol.~21, no.~12, pp. 4695--4708, 2012.

\bibitem{chetouani2021deep}
A.~Chetouani, M.~Quach, G.~Valenzise, and F.~Dufaux, ``Deep learning-based quality assessment of 3d point clouds without reference,'' in \emph{2021 IEEE International Conference on Multimedia \& Expo Workshops}, 2021, pp. 1--6.

\bibitem{fan2022no}
Y.~Fan, Z.~Zhang, W.~Sun, X.~Min, W.~Lu, T.~Wang, N.~Liu, and G.~Zhai, ``A no-reference quality assessment metric for point cloud based on captured video sequences,'' \emph{arXiv preprint arXiv:2206.05054}, 2022.

\bibitem{he2016deep}
K.~He, X.~Zhang, S.~Ren, and J.~Sun, ``Deep residual learning for image recognition,'' in \emph{Proceedings of the IEEE conference on computer vision and pattern recognition}, 2016, pp. 770--778.

\bibitem{feichtenhofer2019slowfast}
C.~Feichtenhofer, H.~Fan, J.~Malik, and K.~He, ``Slowfast networks for video recognition,'' in \emph{Proceedings of the IEEE/CVF international conference on computer vision}, 2019, pp. 6202--6211.

\bibitem{deng2009imagenet}
J.~Deng, W.~Dong, R.~Socher, L.-J. Li, K.~Li, and L.~Fei-Fei, ``Imagenet: A large-scale hierarchical image database,'' in \emph{2009 IEEE conference on computer vision and pattern recognition}.\hskip 1em plus 0.5em minus 0.4em\relax Ieee, 2009, pp. 248--255.

\bibitem{kay2017kinetics}
W.~Kay, J.~Carreira, K.~Simonyan, B.~Zhang, C.~Hillier, S.~Vijayanarasimhan, F.~Viola, T.~Green, T.~Back, P.~Natsev \emph{et~al.}, ``The kinetics human action video dataset,'' \emph{arXiv preprint arXiv:1705.06950}, 2017.

\bibitem{kingma2014adam}
D.~P. Kingma and J.~Ba, ``Adam: A method for stochastic optimization,'' \emph{ICLR}, 2014.

\bibitem{hosu2017konstanz}
V.~Hosu, F.~Hahn, M.~Jenadeleh, H.~Lin, H.~Men, T.~Szir{\'a}nyi, S.~Li, and D.~Saupe, ``The konstanz natural video database (konvid-1k),'' in \emph{2017 Ninth international conference on quality of multimedia experience}.\hskip 1em plus 0.5em minus 0.4em\relax IEEE, 2017, pp. 1--6.

\bibitem{koniq10k}
V.~{Hosu}, H.~{Lin}, T.~{Sziranyi}, and D.~{Saupe}, ``Koniq-10k: An ecologically valid database for deep learning of blind image quality assessment,'' \emph{IEEE Transactions on Image Processing}, vol.~29, pp. 4041--4056, 2020.

\bibitem{statistic-test}
H.~Sheikh, M.~Sabir, and A.~Bovik, ``A statistical evaluation of recent full reference image quality assessment algorithms,'' \emph{IEEE Transactions on Image Processing}, vol.~15, no.~11, pp. 3440--3451, 2006.

\bibitem{simonyan2014very}
K.~Simonyan and A.~Zisserman, ``Very deep convolutional networks for large-scale image recognition,'' \emph{arXiv preprint arXiv:1409.1556}, 2014.

\bibitem{sandler2018mobilenetv2}
M.~Sandler, A.~Howard, M.~Zhu, A.~Zhmoginov, and L.-C. Chen, ``Mobilenetv2: Inverted residuals and linear bottlenecks,'' in \emph{Proceedings of the IEEE conference on computer vision and pattern recognition}, 2018, pp. 4510--4520.

\bibitem{feichtenhofer2020x3d}
C.~Feichtenhofer, ``X3d: Expanding architectures for efficient video recognition,'' in \emph{Proceedings of the IEEE/CVF Conference on Computer Vision and Pattern Recognition}, 2020, pp. 203--213.

\end{thebibliography}

\end{document}